\documentclass[preprint,12pt]{elsarticle}

\usepackage{amssymb}
\usepackage{amsmath}
\usepackage{float} 
\usepackage{comment}
\usepackage{hyperref}
\begin{document}

\begin{frontmatter}



\title{iCost: A Novel Instance-Complexity-Based Cost-Sensitive Learning Framework}





\author[1]{Asif Newaz\corref{cor1}}

\ead{eee.asifnewaz@iut-dhaka.edu}
\cortext[cor1]{Corresponding author}

\author[2]{Asif Ur Rahman Adib}
\ead{asif-ur-rahman@iut-dhaka.edu}

\author[3]{Taskeed Jabid}
\ead{taskeed@ewubd.edu}

\address[1]{Department of Electrical and Electronic Engineering, Islamic University of Technology (IUT), Gazipur, Bangladesh}

\address[2]{Department of Electrical and Electronic Engineering, Ahsanullah University of Science and Technology (AUST), Dhaka, Bangladesh}

\address[3]{Department of Computer Science and Engineering, East West University, Dhaka, Bangladesh}

\begin{abstract}
Class imbalance poses a significant challenge in classification tasks, often causing standard learning algorithms to become biased toward the majority class. Cost-sensitive learning (CSL) addresses this issue by assigning higher penalties to minority-class misclassifications. However, conventional CSL typically applies a uniform penalty to all minority-class instances, ignoring the fact that minority samples may differ substantially in terms of local safety, overlap, boundary ambiguity, and outlier-like behavior. Uniform penalization can therefore introduce undue bias, increasing the number of misclassifications.

In this study, we propose iCost, an instance-complexity-aware CSL framework that assigns adaptive penalties to minority-class samples according to their estimated learning difficulty. This fine-grained penalization strategy ensures fairer weighting, reduces unwarranted bias, and improves overall classification performance. Two complementary complexity estimation strategies are introduced: Neighbor-iCost, based on local neighborhood composition, and Gini-iCost, based on Gini-impurity-based feature-space partitioning. Extensive experiments on 65 binary and 10 multiclass imbalanced datasets show that iCost outperforms conventional CSL by a clear margin and remains highly competitive with widely used resampling methods. To support reproducibility and practical adoption, the proposed algorithm has been released as a scikit-learn-compatible Python package through PyPI.

This work offers a fresh perspective on imbalanced learning by integrating instance-level data complexity into the learning process, opening new avenues for developing adaptive, complexity-aware strategies for imbalanced classification.

\end{abstract}





\end{frontmatter}


\section{Introduction}

Class imbalance occurs when the number of samples differs substantially across classes, with one or more majority classes significantly outnumbering the minority class(es) \cite{branco2016survey}. This issue is common in many real-world applications, including medical diagnosis, fraud detection, spam detection, and fault detection \cite{haixiang2017learning}. In such scenarios, standard machine learning (ML) algorithms often become biased toward the majority class, as they are typically optimized to maximize overall accuracy. For example, in a medical dataset containing 95\% healthy and 5\% diseased patients, a classifier that always predicts the healthy class would achieve 95\% accuracy, while completely failing to identify diseased cases. Such behavior is particularly problematic in high-stakes applications, where misclassifying minority-class instances may lead to serious consequences \cite{newaz2022intelligent}.

To address this issue, imbalanced learning techniques have been developed to improve the recognition of underrepresented yet often critical minority-class instances. These methods aim to reduce majority-class bias and improve generalization across all classes. Over the years, a wide range of approaches has been proposed for imbalanced classification \cite{fernandez2018learning}. These techniques can be broadly categorized into two groups: data-level approaches and algorithm-level approaches.

Data-level approaches modify the training distribution by either generating synthetic minority-class samples or removing selected majority-class samples \cite{susan2021balancing}. This process is commonly referred to as data resampling. Although such methods are commonly used to rebalance class distributions, recent studies suggest that simply balancing the number of samples may be insufficient; reducing class overlap and other data-intrinsic difficulties is often equally important for improving classification performance \cite{santos2023unifying}.

In contrast, algorithm-level approaches address imbalance by modifying the learning process itself. Cost-sensitive learning (CSL) incorporates unequal misclassification costs into the objective function so that errors on minority-class instances receive greater emphasis \cite{elkan2001foundations}. By assigning higher penalties to minority-class misclassifications, CSL shifts the learning bias away from the majority class and encourages the classifier to better recognize underrepresented cases. 

Resampling-based methods can become computationally expensive for large datasets, particularly when synthetic samples are generated or repeated cleaning procedures are applied \cite{krawczyk2016learning}. CSL avoids directly modifying the training distribution and can therefore be more convenient in large-scale settings. However, since CSL does not introduce new minority-class information, its performance may be limited when the data are extremely skewed and only a very small number of minority-class samples are available. Both data-level and algorithm-level approaches have been widely applied in real-world imbalanced classification problems \cite{shen2020dynamic, zhang2020machinery, newaz2021diagnosis, khairy2024effect}.

This study focuses on CSL, where a specific penalty is added to the misclassifications of the minority-class instances. In a conventional error-driven classification setting, misclassification can be viewed through the 0--1 loss perspective, where a correct prediction receives a loss of 0 and an incorrect prediction receives a loss of 1. Under this formulation, all errors are treated equally, regardless of whether they occur on majority- or minority-class instances. Although practical classifiers may optimize different surrogate losses, such as log loss, hinge loss, or impurity-based criteria, the standard learning objective does not explicitly differentiate between majority- and minority-class misclassifications unless class- or instance-level weights are introduced. This becomes problematic under skewed class distributions, where a classifier can achieve high overall accuracy while failing to correctly identify minority-class instances. To avoid this, the idea of the cost-driven classifier is introduced, where asymmetric misclassification cost is utilized. Assigning a higher misclassification cost to the minority-class instances compared to the majority-class forces the algorithm to put more priority on learning those instances correctly, reversing the bias. This approach has been incorporated into several widely used ML libraries, including scikit-learn and XGBoost \cite{scikit-learn, Chen-xg}. 

\begin{table} [b]
    \centering
    \caption{Cost Matrix}
    \begin{tabular}{|l|l|l|l|}
    \hline
        \textbf{} & \textbf{Predicted True} & \textbf{Predicted False} & \textbf{} \\ \hline
        \textbf{Actual True} & 0 & \(C_r\) & \textbf{Minority Class} \\ \hline
        \textbf{Actual False} & \(C_p\) & 0 & \textbf{Majority Class} \\ \hline
    \end{tabular}
    \label{cost}
\end{table}

CSL is implemented through a cost matrix, as shown in Table~\ref{cost}. Here, \(C_r\) and \(C_p\) denote the penalties associated with misclassifying minority-class and majority-class instances, respectively. In the error-driven design, both \(C_r\) and \(C_p\) are set to 1. In the cost-driven setting, \(C_p\) is set to 1, while \(C_r\) is assigned a higher value to increase the model's sensitivity toward the minority class. This improves the recall/sensitivity score. However, increasing \(C_r\) also introduces a trade-off: while it may improve recall, excessive penalization can reduce specificity by increasing the number of majority-class misclassifications. In practice, the minority-class penalty is often selected heuristically or optimized using search-based methods to maintain a good balance between sensitivity and specificity. In the scikit-learn implementation class weights are often derived from the class frequencies and are therefore closely related to the imbalance ratio.

\begin{figure}
\centering
\includegraphics[width=0.95\linewidth]{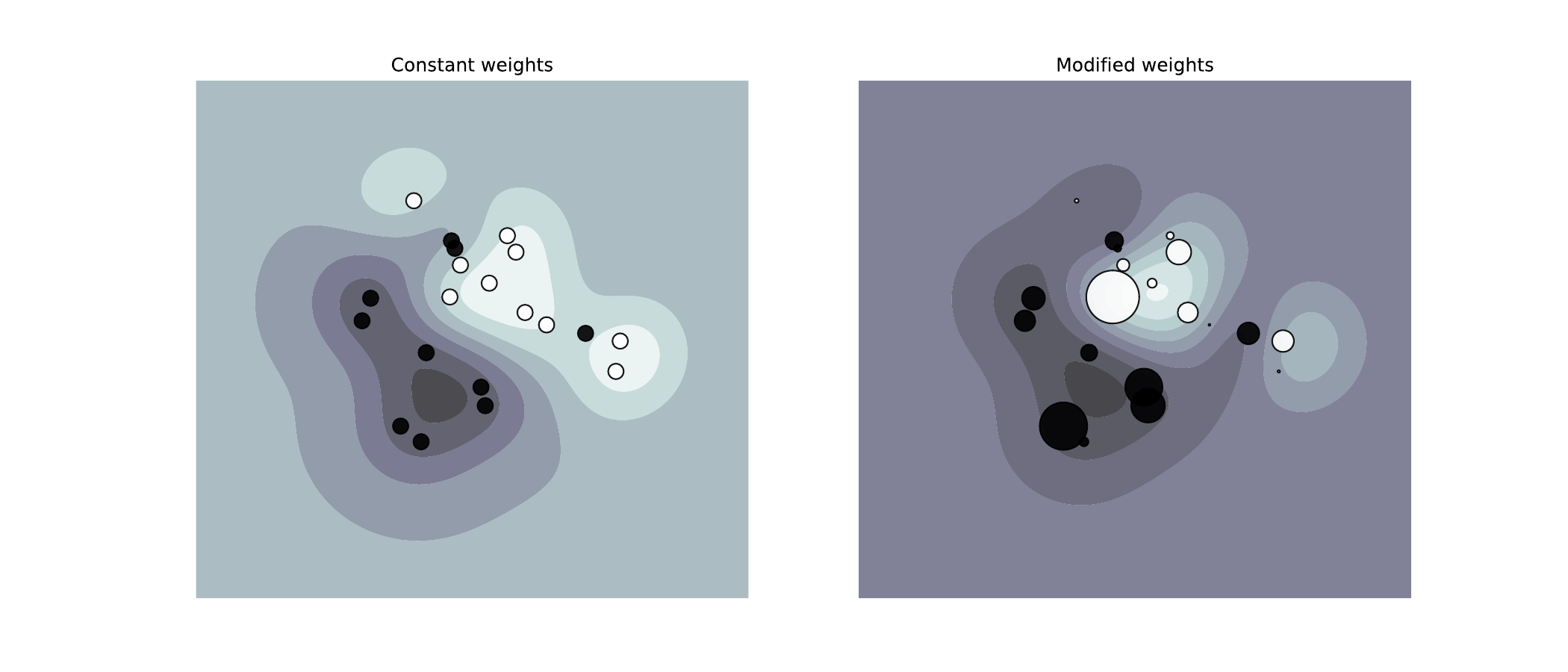}
\caption{Effect of modifying weights on the decision boundary. Here, the size of the points is proportional to their weight.}
\label{wt}
\end{figure}

The effect of instance weighting on the decision boundary of a Support Vector Machine (SVM) classifier is illustrated in Fig.~\ref{wt}. Assigning a higher weight to selected instances forces the classifier to prioritize their correct classification, thereby shifting or deforming the original decision boundary. However, this adjustment may also cause nearby lower-weighted instances to be misclassified, as shown in the figure. Therefore, instance weights must be assigned carefully to avoid excessive boundary distortion and unintended misclassifications.

In the case of the traditional CSL approach, the same cost value is applied to all the minority-class instances indiscriminately, regardless of their individual characteristics \cite{fernandez2018cost}. This creates a major flaw: not all minority-class instances are equally difficult to classify. Samples located near decision boundaries or within overlapping regions are more likely to be misclassified than those located in clearly separable regions. Therefore, hard-to-learn minority instances should receive stronger penalization, while relatively easy instances should not be overemphasized unnecessarily. For example, in a dataset with an IR of 100, a conventional CS strategy may assign all minority-class instances a penalty 100 times larger than that of majority-class instances. While this can improve minority-class sensitivity, it may also excessively distort the decision boundary, especially when well-separated minority samples are over-penalized. Such overcompensation can increase majority-class misclassifications, reduce specificity, and degrade generalization on unseen data.

To address this limitation, penalties should be assigned according to instance-level classification difficulty rather than class membership alone. In this study, we propose iCost, an Instance Complexity-based Cost-sensitive Learning framework, which incorporates this principle to achieve more balanced and context-aware penalization in imbalanced classification. A key challenge is the reliable estimation of individual instance complexity. Without an appropriate measure of difficulty, the assigned penalties may become arbitrary, limiting the effectiveness of the learning process. Among different data-intrinsic difficulty factors, class overlap is widely recognized as one of the most critical factors affecting classifier performance \cite{vuttipittayamongkol2020overlap}. In addition, noisy samples and small disjuncts can further complicate the learning task by introducing ambiguous or poorly represented regions in the feature space \cite{dudjak2021empirical}.

\begin{figure}
\centering
\includegraphics[width=0.95\linewidth]{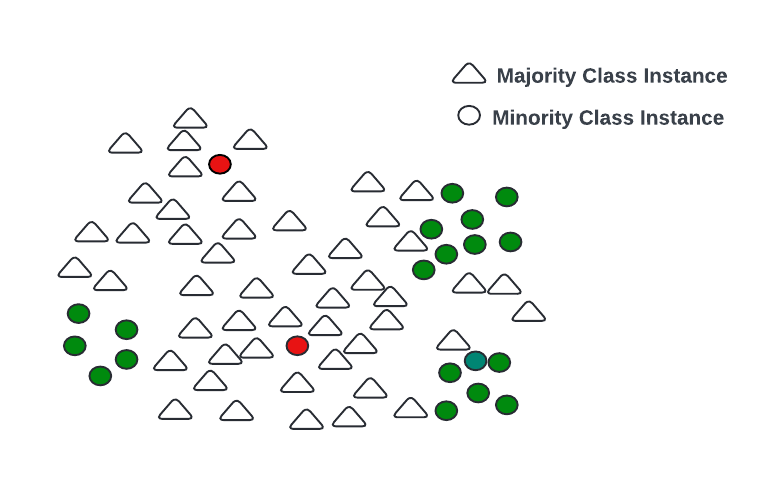}
\caption{Presence of noisy instance in the data (red circle marks the noisy minority-class instances). It also illustrates small disjuncts in the data.}
\label{noise}
\end{figure}

\begin{figure} 
\centering
\includegraphics[width=0.95\linewidth]{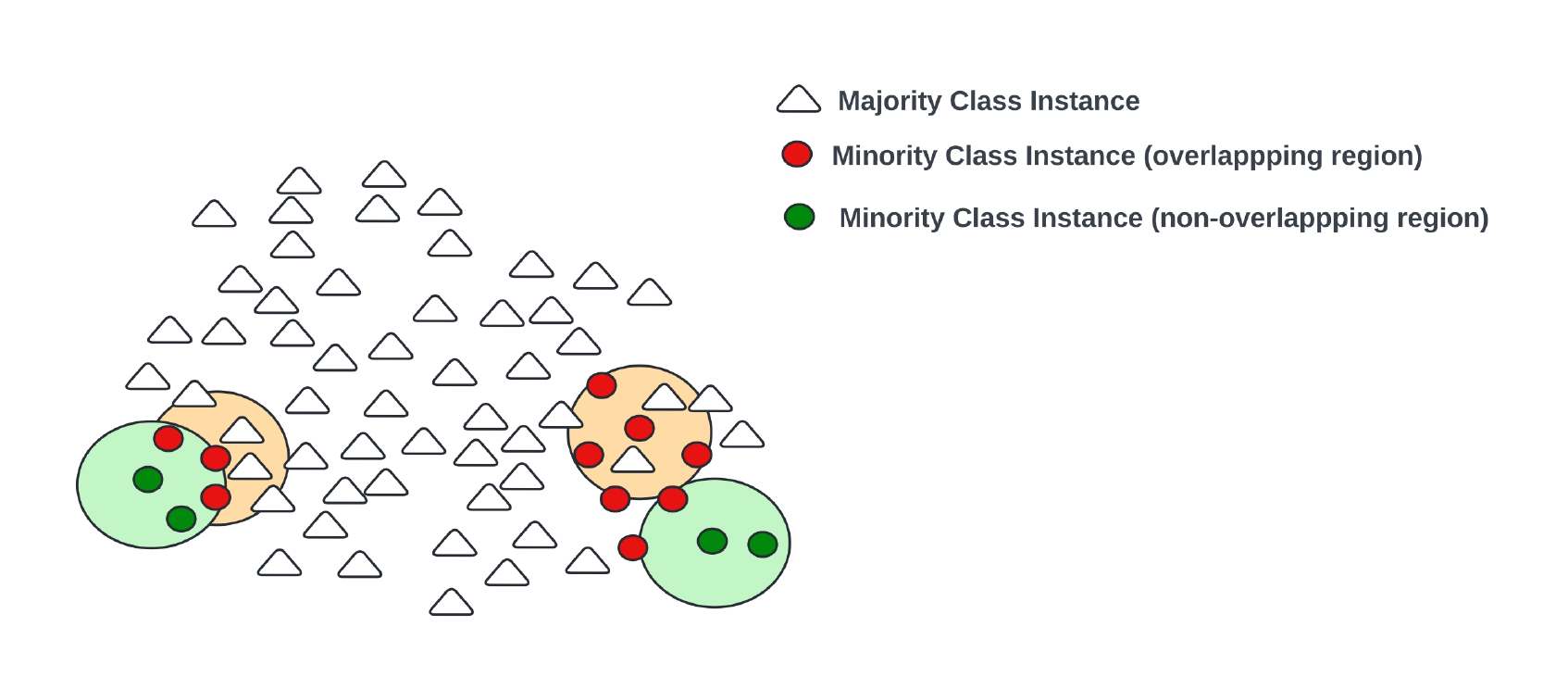}
\caption{Class overlapping between opposite class instances}
\label{overlap}
\end{figure}

These issues are illustrated in Fig.~\ref{noise} and Fig.~\ref{overlap}. In Fig.~\ref{noise}, two minority-class instances are completely surrounded by majority-class samples and are located far from other minority-class instances. Such samples are likely to be noisy or outlying observations. However, under a traditional CS strategy, these instances would still receive the same high penalty as other minority-class samples, potentially forcing the classifier to overfit to unreliable points and increasing misclassifications among nearby majority-class instances. In contrast, Fig.~\ref{overlap} shows minority-class instances located in overlapping regions, where they are surrounded by opposite-class samples and are therefore more difficult to classify. These instances are more informative for shaping the decision boundary and should receive stronger penalization than clearly separable instances. The proposed iCost framework incorporates these considerations by defining instance complexity based on local class composition and feature-space partitioning, considering structural proximity to opposite-class samples.

Researchers have proposed various ways of identifying overlapping and difficult regions in imbalanced datasets. Santos et al. presented a taxonomy of class overlap measures, demonstrating different ways of representing class overlap \cite{santos2022joint}. For example, feature-based measures, such as the maximum Fisher’s discriminant ratio, estimate the extent to which individual features can separate the classes. Other approaches examine overlap through local neighborhoods, structural properties, clustering behavior, or partitioned regions of the feature space.

In this study, we define instance complexity using two complementary methodologies. The first is a neighborhood-based approach, which estimates the difficulty of a minority-class instance from the class composition of its local neighborhood. This allows the framework to distinguish relatively boundary/overlapping samples from easy or outlier-like samples. The second is a Gini-impurity-based feature-space partitioning approach, where a shallow decision tree is used as a probe model to partition the feature space. The class distribution within the leaf node containing a minority-class instance is then used to estimate its regional ambiguity. Minority instances located in highly mixed leaves are considered more complex, while those located in mostly minority-dominated leaves are treated as easier samples. This provides a feature-partition-based view of instance complexity that complements the local neighborhood-based view. Together, these two approaches allow iCost to account for both local class composition and regional class impurity when assigning instance-level penalties. Thus, the proposed framework can emphasize informative overlapping samples while limiting the influence of clearly separable or potentially noisy minority-class instances.

In the conventional CSL approach, class-level penalization alone cannot distinguish between minority instances that are easy, ambiguous, or potentially unreliable. This leads to the central research question of this study: how can we design a cost-sensitive learning framework that adaptively penalizes instances based on their classification complexity, rather than relying solely on class labels? To address this question, we propose iCost, an Instance Complexity-based Cost-sensitive Learning framework. The proposed framework estimates the difficulty of minority-class instances and assigns instance-specific penalties accordingly. Instead of uniformly increasing the weight of all minority samples, iCost emphasizes informative overlapping and boundary samples while reducing unnecessary over-penalization of easy-to-classify samples. It also carefully identifies outlier-like samples and reduces their penalty to avoid misclassification of nearby majority-class instances. This produces a more context-aware cost assignment strategy for imbalanced classification tasks, resulting in better generalization and improved classification performance. Extensive experiments on 75 imbalanced datasets show that our proposed approach improves over traditional CSL across multiple classifiers, providing consistent gains in performance measures.


\vspace{\baselineskip}

The main contributions of this work are as follows.

\begin{itemize}
    \item We identify a key limitation of conventional CSL strategies: the use of uniform class-level penalties for all minority-class instances.
    \item We propose a novel, instance complexity-based CSL framework that assigns adaptive penalties to minority-class instances based on their estimated difficulty, enabling more effective learning in imbalanced settings.
    \item We introduce two complementary strategies for estimating instance complexity. The first uses neighborhood composition and the second uses Gini-impurity-based feature-space partitioning.
    \item We define a complexity-aware penalty assignment strategy in which informative boundary and overlapping samples receive higher penalties, while clearly separable or outlier-like samples receive lower penalties to reduce unnecessary boundary distortion and overfitting.
    \item We conduct extensive experiments on 65 binary imbalanced datasets using five different classifiers to evaluate the effectiveness of the proposed framework against standard CSL and resampling-based methods.
    \item We further extend the proposed framework to multiclass imbalanced classification and evaluate its performance on 10 multiclass datasets.
    \item To support reproducibility and practical adoption, the full implementation is made publicly available through GitHub and released as a scikit-learn-compatible Python package on PyPI.
    
\end{itemize}

To the best of our knowledge, the integration of instance-level complexity into a general CSL framework remains underexplored in the existing literature. This work introduces a more flexible and complexity-aware direction for imbalanced classification, providing a foundation for future research on adaptive learning strategies that better align with the intrinsic structure of the data.

The rest of the article is organized as follows. Related work is discussed in Section 2. The proposed methodology is presented in Section 3. The experimental design is described in Section 4. Results and discussion are presented in Section 5. Finally, Section 6 concludes the manuscript.


\section{Related Works}

\subsection{General Approaches to Imbalanced Learning}

A wide range of techniques have been developed to address the imbalanced classification problem, which can be broadly grouped into data-level approaches, algorithm-level approaches, and hybrid approaches \cite{leevy2018survey, rezvani2023broad}. Data-level methods modify the training distribution before model learning. Oversampling techniques increase the representation of the minority class by duplicating or generating minority-class samples. Common examples include random oversampling (ROS), SMOTE \cite{chawla2002smote}, and several SMOTE-based variants, such as Borderline-SMOTE (BL-SMOTE) \cite{han2005borderline}, Geometric-SMOTE \cite{douzas2017geometric}, and SMOTE-IPF \cite{saez2015smote}. In contrast, undersampling techniques reduce the dominance of the majority class by removing selected majority-class samples. Popular examples include random undersampling (RUS), instance hardness thresholding (IHT) \cite{smith2014instance}, and edited nearest neighbors (ENN). It has been reported that algorithm-level CS approaches generally outperform undersampling methods but cannot
surpass oversampling techniques such as SMOTE or ADASYN \cite{mohosheu2024comprehensive}.

Hybrid strategies combine different mechanisms, such as oversampling followed by cleaning or cost-sensitive training. Examples include SMOTE-ENN, SMOTE-Tomek, and combinations of resampling with cost-sensitive classifiers \cite{thai2010cost, newaz2023predicting}. These approaches can improve minority-class recognition by reshaping the training distribution or modifying the learning objective. However, most existing methods primarily focus on balancing the class distribution rather than explicitly modeling the intrinsic difficulty of individual samples. As a result, data-intrinsic factors such as class overlap or local ambiguity remain insufficiently addressed \cite{krawczyk2016learning}.

\subsection{Handling Data Difficulty Factors}

Recent studies have emphasized that class imbalance alone does not fully explain the complexity of a classification task. Several data-intrinsic factors can make the learning task more challenging, including class overlap, noisy samples, small disjuncts, rare cases, and borderline instances \cite{dudjak2021empirical}. Among these factors, class overlap is often considered one of the most influential, as samples from different classes may occupy similar regions of the feature space, making the decision boundary difficult to define \cite{vuttipittayamongkol2021class}. Therefore, improving performance on imbalanced datasets requires not only correcting class frequencies but also considering the structural and local characteristics of the data.

Several approaches have been proposed to address these difficulty factors \cite{Stefanowski2016, koziarski2019radial}. Many of them are based on data-level preprocessing, such as removing noisy or borderline majority-class instances, selectively oversampling informative minority-class samples, or applying cleaning procedures after synthetic sample generation. Other methods use local data difficulty to guide ensemble selection or to determine which samples should be retained, removed, or oversampled. These approaches demonstrate the importance of incorporating data complexity into imbalanced learning.

However, most existing methods that explicitly consider data difficulty factors operate by modifying the training distribution through oversampling or undersampling. In contrast, very few studies have explored how such factors can be incorporated directly into the CSL process. This leaves an important gap: data complexity is widely recognized as critical in imbalanced learning, yet its explicit integration into general CSL frameworks remains limited. The proposed iCost framework addresses this gap by incorporating instance-level difficulty into the penalty assignment process without altering the original data distribution.

\subsection{Cost-Sensitive Learning Frameworks}

Instead of changing the class distribution, CSL modifies the learning objective by assigning different costs to different types of misclassification. In binary imbalanced classification, errors on minority-class instances are typically assigned a higher penalty than errors on majority-class instances. This encourages the classifier to pay greater attention to the underrepresented class during training \cite{elkan2001foundations}. Several conventional classifiers have been adapted to the cost-sensitive setting, including SVMs, Artificial Neural Networks (ANNs), Decision Trees (DTs), Random Forests (RFs), and boosting-based methods \cite{iranmehr2019cost, zhou2005training}. These adaptations are often implemented by introducing class weights into the loss function, impurity criterion, margin objective, or ensemble weighting mechanism. Popular ML libraries, such as scikit-learn and XGBoost, provide built-in support for class-weighted training, making CSL practically convenient for many applications \cite{scikit-learn, Chen-xg}.

Despite its practical usefulness, traditional CSL usually relies on a predefined class-level cost matrix which is typically set according to the IR of the dataset. While this can reduce majority-class bias, it treats all minority-class samples as equally important and equally difficult to classify. This assumption is often unrealistic in complex imbalanced datasets, where minority samples may differ substantially in terms of local ambiguity, overlap, noise, and separability. As a result, conventional CSL may over-penalize easy or unreliable minority samples while failing to selectively emphasize the most informative boundary samples. This limitation motivates the development of instance-level cost-sensitive strategies that adapt penalties according to sample-specific difficulty.

\subsection{Algorithm-Level Strategies}

Although most overlap- and noise-aware methods rely on data-level preprocessing, some studies have attempted to address these issues through algorithm-level modifications. Lee and Kim proposed an Overlap Sensitive Margin (OSM) classifier for imbalanced and overlapping data \cite{LEE201872}. Their approach combines a modified Fuzzy SVM with a k-Nearest Neighbor (kNN) mechanism to distinguish between soft-overlap and hard-overlap regions. Similarly, Natarajan et al. investigated cost-sensitive classification under class-conditional label noise by adapting surrogate loss functions such as logistic and hinge losses \cite{natarajan2018cost}. Example-dependent CSL has also been studied, particularly in application-specific domains such as credit scoring, where the cost associated with an instance may depend on external information such as financial risk or customer profile \cite{bahnsen2014example, zelenkov2019example}. However, these approaches usually rely on domain-defined cost information and may not generalize well to arbitrary imbalanced datasets.

\subsection{Summary and Motivation}

The reviewed literature shows that data difficulty factors are widely recognized as important in imbalanced classification, but they are more commonly addressed through resampling, instance selection, or data cleaning. In conventional CSL, penalty assignment still largely remains class-level, with limited consideration of how individual minority-class samples differ in difficulty. Existing algorithm-level extensions address related problems, but a general CSL framework that derives penalties from data-intrinsic instance complexity remains limited. This gap motivates the proposed iCost framework, which incorporates instance-level difficulty into the cost assignment process while preserving compatibility with standard ML classifiers.

\section{Methodology}

This section presents the proposed iCost framework. We first describe the conventional cost-sensitive learning formulation and then mathematically show how instance-complexity-based weighting can be incorporated into the CSL framework. Next, we discuss the estimation of instance complexity and present two complementary strategies. We then describe the complexity-aware penalty assignment mechanism, followed by the implementation details and the extension of the proposed framework to multiclass classification.

\subsection{Cost-Sensitive Classifier}

In classification algorithms, a cost function is utilized for the learning process. The goal is to minimize the total cost associated with the predictions. Optimization algorithms such as stochastic gradient descent are utilized to reach the optimal point and the model parameters are determined through this process. For example, in binary classification for the LR classifier, log loss is calculated in the following way:

\begin{equation}
\text{Log Loss} = - \frac{1}{N} \sum_{i=1}^{N} \left[ y_i \log(h_\theta(x_i)) + (1 - y_i) \log(1 - h_\theta(x_i)) \right]
\end{equation}

Here, 
\begin{itemize}
    \item N is the total number of samples
    \item \( y_i \) is the true label of the ith sample (0 or 1)
    \item \( h_\theta(x_i)\) is the predicted probability that the ith sample belongs to class 1, as calculated by the classifier.
    \item \(\theta\) is the model parameters.
\end{itemize}

When the predicted probability assigned to the true class is high, the log loss becomes small; and when it is low, the loss increases. In the conventional formulation, no distinction is made between positive- and negative-class errors, and all samples contribute to the loss according to the same weighting scheme. To make the classifier cost-sensitive, the log-loss formulation can be modified by introducing different class weights.  

\begin{equation}
  \text{CS Log Loss} = - \frac{1}{N} \sum_{i=1}^{N} \left[ C_1 \cdot y_i \log(h_\theta(x_i)) + C_0 \cdot (1 - y_i) \log(1 - h_\theta(x_i)) \right]  
\end{equation}

Here, \(C_1\) and \(C_0\) denote the penalties of misclassifying positive and negative classes, respectively. Usually, the \(C_0\) value is kept fixed at 1 while the \(C_1\) is assigned a different weight (typically, the IR of the dataset). This allows the algorithm to put more emphasis on correctly predicting the minority-class (positive) instances. Thus, modifying the learning process.

While this approach certainly improves the model's ability to correctly identify the minority-class instances, this usually comes at the cost of many misclassifications of the majority-class instances as they are assigned a lower weight. The trade-off becomes a defining factor in determining how many misclassifications of the majority-class instances can be justified in exchange for a few correct predictions of minority-class instances. Although in certain applications, such as medical diagnosis or fault detection, correctly identifying the positive cases is crucial, too many false positives reduce the reliability of the prediction framework.

\subsection{Proposed Instance-Complexity Based Cost-Sensitive Classifier}

The proposed iCost framework introduces instance-level penalty assignment into CSL. Instead of applying a uniform minority-class cost, iCost first estimates the complexity of each minority-class instance and then assigns a penalty according to its difficulty. Minority-class samples located near overlapping or boundary regions receive higher penalties, while clearly separable samples that are away from the decision boundary receive lower penalties. Penalties for outlier or noisy samples are set even lower. This allows the classifier to focus on useful samples without unnecessarily distorting the decision boundary around easy or unreliable instances.

\vspace{\baselineskip}

Mathematically, it can be represented as follows:  

\begin{equation}
    \text{Modified CS Log Loss} = - \frac{1}{N} \sum_{i=1}^{N} \left[ C_{j(i)} \cdot y_i \log(h_\theta(x_i)) + (1 - y_i) \log(1 - h_\theta(x_i)) \right]
\end{equation}

Here, \(C_j(i)\) is the cost for the category j that instance i belongs to. One thing to consider here is that only the minority-class instances (positive cases) are categorized and penalized with different weights. The majority-class instances are assigned equal weights, in line with the traditional cost-sensitive approach. While majority-class instances can also be categorized in a similar fashion, that would increase the complexity of the algorithm and is not explored in this article. 

This concept is equally applicable to other classification algorithms, such as SVM or DTs. For the SVM classifier, the cost (objective) function aims to find a hyperplane that maximizes the margin between classes while penalizing misclassified points. For a linear SVM, the objective function can be expressed as follows:

\begin{equation}
    \text{Minimize} \quad \frac{1}{2} ||\mathbf{w}||^2 + C \sum_{i=1}^{N} \max(0, 1 - y_i (\mathbf{w} \cdot \mathbf{x}_i + b))
\end{equation}

Here,
\begin{itemize}
    \item w is the weight vector defining the hyperplane.
    \item \(||w||^2\) is the regularization term that controls the margin width.
    \item C is the penalty parameter that controls the trade-off between maximizing the margin and minimizing classification errors.
    \item \(y_i\) is the true label for instance i.
    \item \(x_i\) is the feature vector for instance i.
    \item b is the bias term.
    \item \(max(0, 1 - y_i (\mathbf{w} \cdot \mathbf{x}_i + b)\)) is the hinge loss, penalizing instances on the wrong side of the margin.
\end{itemize}

The function can be modified to better handle imbalanced data by penalizing errors on minority-class instances more heavily. The objective function for the CS-SVM classifier can be expressed as follows:

\begin{equation}
  \text{Minimize} \quad \frac{1}{2} ||\mathbf{w}||^2 + C_+ \sum_{i \in \mathcal{P}} \max(0, 1 - y_i (\mathbf{w} \cdot \mathbf{x}_i + b)) + C_- \sum_{i \in \mathcal{N}} \max(0, 1 - y_i (\mathbf{w} \cdot \mathbf{x}_i + b))  
\end{equation}

Here, 

\begin{itemize}
    \item \(C_+\) is the penalty parameter for the positive class (often the minority class).
    \item \(C_-\) is the penalty parameter for the negative class (often the majority class).
    \item \(\mathcal{P}\) and \(\mathcal{N}\) represent the sets of positive and negative instances, respectively.
\end{itemize}

The proposed instance-based weighting formula can be incorporated into the equation in the following way: 

\begin{equation}
    \text{Minimize} \quad \frac{1}{2} ||\mathbf{w}||^2 + C \sum_{i \in \mathcal{M}} \max(0, 1 - y_i (\mathbf{w} \cdot \mathbf{x}_i + b)) + \sum_{j=1}^{k} C_j \sum_{i \in \mathcal{C}_j} \max(0, 1 - y_i (\mathbf{w} \cdot \mathbf{x}_i + b))
\end{equation}

Here, 
\begin{itemize}
    \item C is the penalty parameter for all majority-class instances, represented by set \(\mathcal{M}\).
    \item \(C_j\) is the penalty parameter specific to category j within the minority class. 
    \item k is the number of categories in the minority class.
\end{itemize}

This way, the CSL framework of different classifiers can be adjusted to incorporate the instance-complexity-based weighting mechanism into the algorithm.

\subsection{Instance Complexity Estimation}

The first stage of the proposed iCost framework is to estimate the complexity of minority-class instances. Two complementary strategies are used in this study for this purpose. The first strategy is a distance-based approach that uses local neighborhood composition to identify where a minority-class sample is located relative to opposite-class instances. The second strategy uses impurity-based feature-space partitioning to estimate the regional ambiguity of a minority-class sample. These two strategies provide two different ways of defining instance difficulty.

\subsubsection{Neighborhood Search}

The neighborhood-based strategy estimates the difficulty of a minority-class instance from the labels of its nearest neighbors. The underlying intuition is that a minority sample surrounded mostly by minority-class samples is relatively easy to classify, whereas a minority sample surrounded by many majority-class samples is more ambiguous and likely to be located near an overlapping or boundary region.

For each minority-class instance \(x_i\), the k-nearest neighbors (k=5) are identified using Euclidean distance (default) in the feature space. Let \(m_i\) denote the number of majority-class samples among the k-nearest neighbors of \(x_i\). Based on \(m_i\), each minority-class instance is assigned to one of the four following categories: 

\begin{itemize}
\item \textbf{Pure:} \(m_i = 0\). The minority instance is surrounded entirely by minority-class samples and is considered relatively easy to classify.
\item \textbf{Safe:} $m_i \in \{1,2\}$. The instance has a small number of majority-class neighbors and may require moderate attention during training.
\item \textbf{Border:} \(m_i \epsilon\) \{3, 4\}. The instance is located in a locally mixed region and is close to the decision boundary or overlap region. These samples would be misclassified by the KNN classification rule.
\item \textbf{Outlier/Noisy:} \(m_i = 5\). The instance is completely surrounded by majority-class samples and may represent a noisy, isolated, or highly ambiguous minority sample.
\end{itemize}



\begin{figure}
\centering
\includegraphics[width=0.95\linewidth]{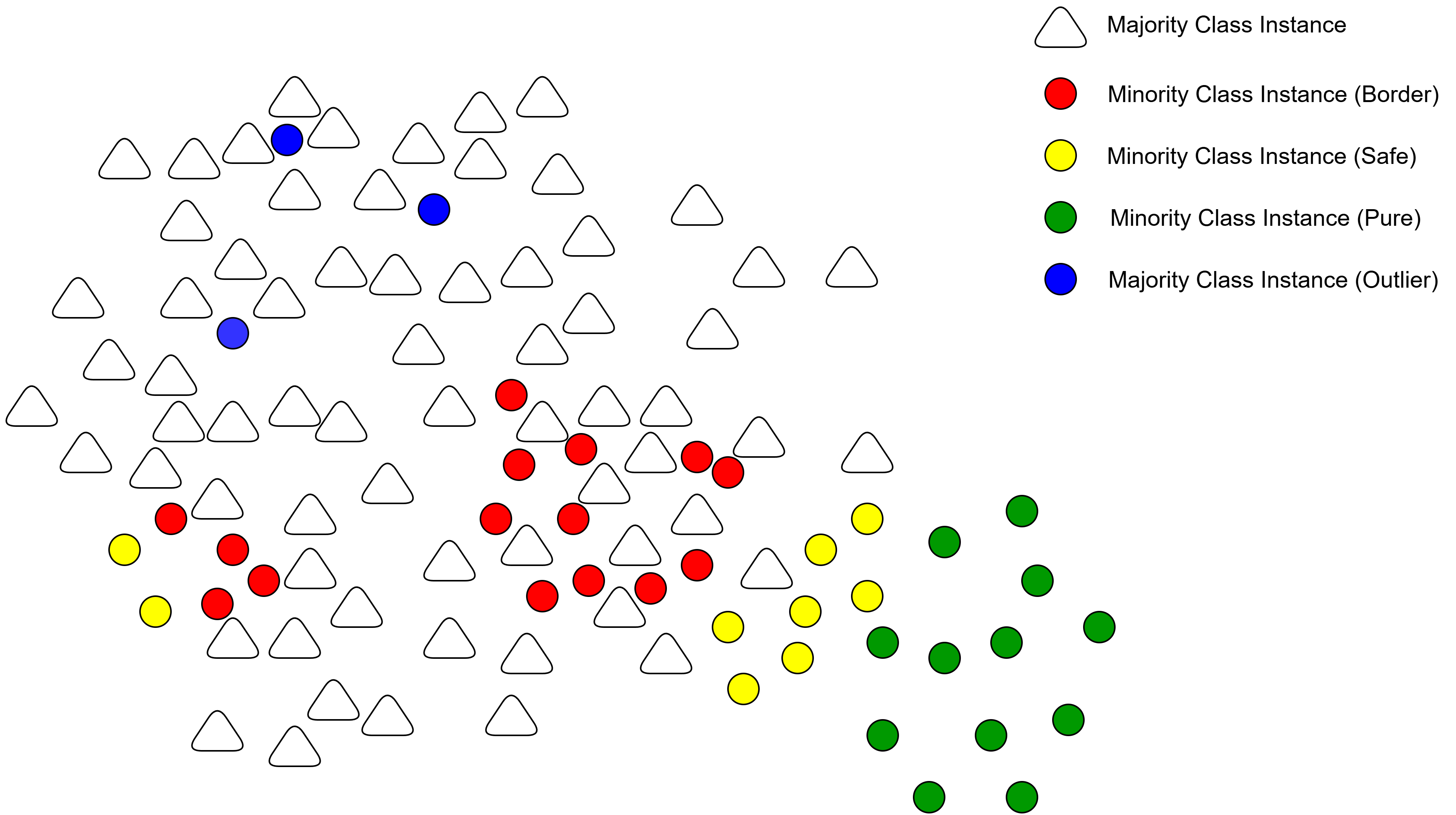}
\caption{Categorization of Minority-class instances}
\label{catg}
\end{figure}

This is illustrated in Fig. \ref{catg}. This categorization allows iCost to distinguish boundary samples from clearly separable and potentially unreliable samples. The categorization formula is flexible and can be easily modified. This option is provided in the implementation of the algorithm.


\subsubsection{Gini Impurity}

This strategy estimates instance complexity using a Gini-impurity-based feature-space partitioning approach. In this method, a shallow decision tree is used as a probe model to partition the training data into a set of leaf nodes. The tree is not used as the final classifier; rather, it is used to identify regions of the feature space where minority- and majority-class samples are either well separated or highly mixed.

\begin{figure}
\centering
\includegraphics[width=0.95\linewidth]{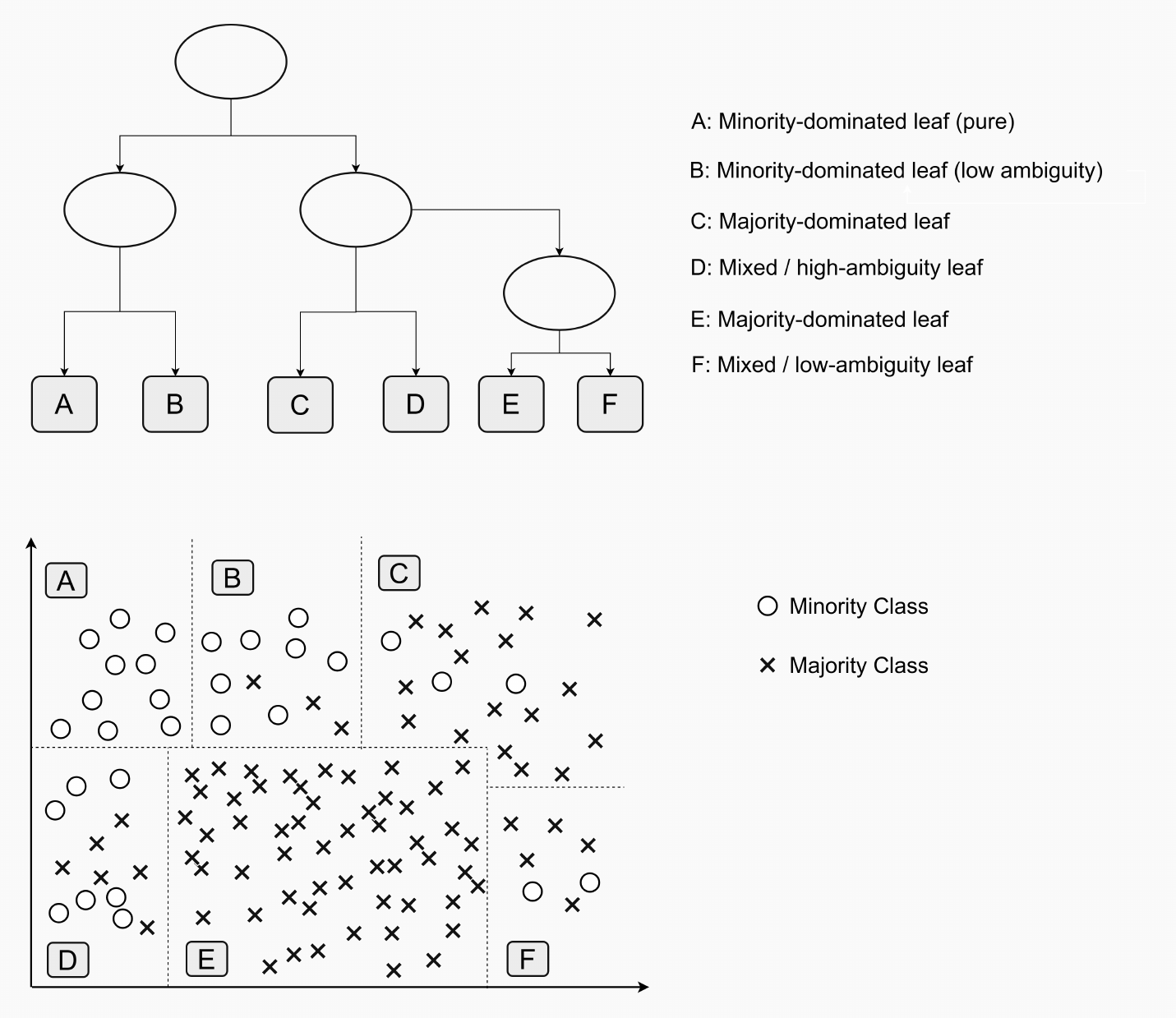}
\caption{Illustration of the Gini-iCost complexity estimation strategy.}
\label{fig:gini_partition}
\end{figure}

Figure~\ref{fig:gini_partition} illustrates the feature-space partitioning view of Gini-iCost. The probe tree divides the input space into several leaf regions with different class compositions. For each minority-class instance, the leaf-level class distribution is used to estimate regional ambiguity. Minority samples in minority-dominated leaves are considered easier, while those in mixed leaves are considered more ambiguous and receive penalties according to the normalized Gini impurity. Minority samples located in majority-dominated leaves are treated as outlier-like or highly isolated samples and receive reduced penalties. Thus, the tree is not used as the final classifier; it is only used to estimate regional complexity for instance-level cost assignment.

After fitting the probe tree on the training data, each minority-class instance is assigned to a leaf node. Let \(L_i\) denote the leaf node containing the minority-class instance \(x_i\). The class distribution inside \(L_i\) is then used to estimate the regional ambiguity around \(x_i\). Let \(n_{min}(L_i)\) and \(n_{maj}(L_i)\) denote the number of minority- and majority-class samples in the leaf node, respectively. The minority-class proportion inside the leaf can be written as:

\begin{equation}
p_{min}(L_i) = \frac{n_{min}(L_i)}{n_{min}(L_i) + n_{maj}(L_i)}.
\end{equation}

Similarly, the majority-class proportion is given by - 

\begin{equation}
p_{maj}(L_i) = 1 - p_{min}(L_i).
\end{equation}

The impurity of the leaf node is calculated using the Gini impurity criterion:

\begin{equation}
G(L_i) = 1 - p_{min}(L_i)^2 - p_{maj}(L_i)^2.
\end{equation}

For binary classification, the maximum Gini impurity is 0.5, which occurs when the two classes are equally represented in a leaf. Therefore, a normalized regional ambiguity score can be defined as - 

\begin{equation}
s_i = 2G(L_i),
\end{equation}

where \(s_i \in [0,1]\). A value close to 0 indicates that the leaf is dominated by one class and is therefore relatively pure, while a value close to 1 indicates a highly mixed region with strong class ambiguity. Based on the minority-class proportion inside the leaf, each minority-class instance is categorized into one of three regional types:

\begin{itemize}
\item \textbf{Minority-dominated leaf:} If \(p_{min}(L_i) \geq \tau_{pure}\), the instance lies in a region mostly occupied by minority-class samples and is considered relatively easy to classify.

\item \textbf{Mixed leaf:} If \(\tau_{out} < p_{min}(L_i) < \tau_{pure}\), the instance lies in a region containing both minority- and majority-class samples. Such instances are considered regionally ambiguous, and their difficulty is determined using the normalized Gini score \(s_i\).

\item \textbf{Majority-dominated leaf:} If \(p_{min}(L_i) \leq \tau_{out}\), the instance lies in a region dominated by majority-class samples. Such samples may represent noisy or highly isolated minority instances and should not be overemphasized during training.

\end{itemize}

In this study, \(\tau_{pure}=0.80\) and \(\tau_{out}=0.20\) are used as default thresholds. Thus, minority samples located in highly mixed leaves are treated as more complex, while samples located in mostly minority-dominated leaves are treated as easier. Samples located in strongly majority-dominated leaves are considered noisy and receive limited emphasis. Thus, this strategy provides a regional view of instance complexity.

\subsection{Complexity-Aware Penalty Assignment}

After estimating the complexity of minority-class instances, the next step is to assign suitable misclassification penalties. In conventional CSL, all minority-class samples receive the same penalty, usually based on the imbalance ratio. In contrast, iCost assigns different penalties to minority-class samples according to their estimated difficulty. 

Let IR denote the imbalance ratio of the dataset, defined as - 

\begin{equation}
IR = \frac{N_{maj}}{N_{min}},
\end{equation}

where \(N_{maj}\) and \(N_{min}\) are the numbers of majority- and minority-class samples, respectively. 

For the neighborhood-based strategy, four penalty factors are used:
\(c_{fp}\), \(c_{fs}\), \(c_{fb}\), and \(c_{fo}\). They denote the penalties for pure, safe, border, and noisy samples, respectively. In this study, we initially use the following default penalty values - 

\begin{equation}
c_{fo}=0.20IR,\quad
c_{fp}=0.50IR,\quad
c_{fs}=0.75IR,\quad
c_{fb}=1.25IR.
\end{equation}


Thus, border or overlapping samples receive the highest penalty because they are considered the most informative for shaping the decision boundary and more likely to be misclassified as majority class. Pure and safe samples receive lower penalties because they are relatively easier to classify. Outlier-like samples receive the lowest penalty to limit their influence and reduce the risk of overfitting to isolated or noisy minority instances.

For the tree-impurity-based strategy, the assigned penalty depends on the class composition and impurity of the leaf node containing the minority-class instance. If the instance falls in a minority-dominated leaf, it is treated as relatively easy and assigned \(c_{fp}\). If it falls in a majority-dominated leaf, it is treated as outlier/noisy and assigned \(c_{fo}\). For mixed leaves, the penalty is interpolated between the safe and border penalties using the normalized ambiguity score \(s_i\). This formulation allows the tree-based strategy to assign gradually increasing penalties to samples located in more ambiguous leaf regions.

\begin{equation}
C_i =
\begin{cases}
c_{fp}, & \text{if } p_{min}(L_i) \geq \tau_{pure},\\
c_{fs} + (c_{fb}-c_{fs})s_i, & \text{if } \tau_{out} < p_{min}(L_i) < \tau_{pure},\\
c_{fo}, & \text{if } p_{min}(L_i) \leq \tau_{out}.
\end{cases}
\end{equation}


The resulting instance-specific penalties are passed to the base classifier as sample weights during training. Majority-class samples are assigned a fixed weight of 1, while minority-class samples receive adaptive weights according to their estimated complexity. The default penalty values are selected to preserve the intended complexity hierarchy while keeping the scale comparable to conventional CSL. Therefore, the IR value is used as the reference point. For instance, pure samples are assigned half of the IR as penalty since these samples are relatively far from the decision boundary surrounded by same-class instances. These values are not intended to be universally optimal; rather, they provide a fixed, interpretable, and reproducible default configuration for evaluating the proposed framework across datasets and classifiers. The effect of varying these penalty values is examined separately in the cost-parameter sensitivity analysis section.



\subsection{Implementation Details}

Given a training set and a base classifier, iCost first identifies the minority and majority classes and computes the IR. It then estimates the complexity of each minority-class instance using either the neighborhood-based or tree-impurity-based strategy. Based on the estimated category or ambiguity score, an instance-specific penalty is assigned to each minority-class sample, while all majority-class samples are assigned a fixed weight of 1. The resulting sample-weight vector is passed to the base classifier during training. Thus, iCost modifies the learning process through adaptive instance-level weighting without changing the original training distribution.

The proposed iCost framework is implemented in Python and designed to be compatible with the scikit-learn API. The framework inherits from \texttt{BaseEstimator} and \texttt{ClassifierMixin}, enabling integration with standard scikit-learn tools such as pipelines, cross-validation, and hyperparameter search. Any base classifier that supports the \texttt{sample\_weight} argument during training can be used with the proposed framework.

The implementation supports three main training modes. The first mode corresponds to the non-cost-sensitive (ncs) classifier, where all samples are assigned equal weights. The second mode corresponds to traditional CSL, where all minority-class instances receive the same class-level penalty. The third mode corresponds to the proposed iCost framework, where minority-class penalties are assigned adaptively using either neighborhood-based or tree-impurity-based instance complexity estimation.

The main input parameters of the implementation are summarized in Table~\ref{tab}. These parameters allow users to select the base classifier, choose the complexity estimation strategy, and control the penalty values assigned to different types of minority-class instances. Unless explicitly specified by the user, the penalty factors are assigned according to the default configuration as reported in Equations 12 and 13. However, penalty factors can also be set manually by supplying the values as a dictionary. For example, the following dictionary illustrates how user-defined cost values can be supplied:

\begin{verbatim}
    cost_factor = {
        "cfo": 0.10 * IR,
        "cfp": 0.50 * IR,
        "cfs": 1.00 * IR,
        "cfb": 1.25 * IR
    }
\end{verbatim}

This allows the framework to be used either with fixed default penalties or with tuned penalty values. In applications where optimal performance is required, these penalty factors can be optimized using grid search, random search, Bayesian optimization, or evolutionary algorithms. In the experimental evaluation of this study, fixed default values are used to provide a consistent and reproducible comparison across datasets.

\begin{table}[!htbp]
\centering
\caption{Main parameters of the proposed iCost implementation.}
\label{tab}
\begin{tabular}{p{0.25\linewidth} p{0.65\linewidth}}
\hline
\textbf{Parameter} & \textbf{Description} \\
\hline
\texttt{base\_classifier} & ML classifier that supports sample-weighted training, such as LR, SVM, DT, RF, or XGBoost. \\

\texttt{method} & The training mode. Supported options are \texttt{ncs}, \texttt{cs}, \texttt{neighbor}, and \texttt{tree}. \\

\texttt{n\_neighbors} & Number of nearest neighbors used in the neighborhood-based strategy. The default value is set to 5. \\

\texttt{cfp, cfs, cfb, cfo} & Penalty factors associated with minority-class categories under 'neighbor' mode. \\

\texttt{tau\_pure, tau\_out} & Threshold used to identify minority and majority-dominated leaves in the 'tree' mode. \\

\texttt{tree\_max\_depth} & Maximum depth of the shallow decision tree used as a probe model in the 'tree' mode. \\

\texttt{min\_samples\_leaf} & Minimum number of samples required in a leaf node of the probe tree. This helps avoid overly small or unstable leaf regions. \\

\hline
\end{tabular}
\end{table}

A simplified example of using the proposed framework is shown below:

\begin{verbatim}
        model = iCost(
            base_classifier = LogisticRegression(),
            method = "neighbor"
        )
        model.fit(X_train, y_train)
        y_pred = model.predict(X_test)
\end{verbatim}

Similarly, the tree-impurity-based strategy can be used by setting the method to \texttt{tree}:

\begin{verbatim}
        model = iCost(
            base_classifier = XGBClassifier(),
            method = "tree",
            tree_max_depth = 3,
            tree_min_samples_leaf = 5,
            tau_pure = 0.75,
            tau_out = 0.1
        )
\end{verbatim}

The framework is released as a scikit-learn-compatible Python package on PyPI (\url{https://pypi.org/project/icost/}), allowing users to install and apply iCost directly within standard ML workflows. 

\subsection{Extension - Multiclass Classification}

The proposed iCost framework can also be extended to multiclass imbalanced classification. Multiclass imbalance is more challenging than binary imbalance because multiple minority and majority classes may exist simultaneously, and the degree of imbalance and overlapping may vary across different class pairs. To handle this setting, we adopt a decomposition-based strategy \cite{galar2011overview}.

In this study, the multiclass problem is transformed into multiple binary classification problems using the one-vs-rest (OvR) scheme. For each class, a separate binary classifier is trained by treating that class as the positive class and all remaining classes as the negative class. The proposed iCost framework is then applied independently within each binary subproblem. In each OvR classifier, the minority and majority groups are identified from the corresponding binary labels, the imbalance ratio is computed, and instance-specific penalties are assigned to the positive-class samples according to their estimated complexity.

This formulation allows the proposed binary instance-weighting mechanism to be used directly in multiclass settings without requiring a separate multiclass cost matrix. Since iCost follows the scikit-learn estimator structure, it can be used as the base estimator within standard decomposition wrappers such as \texttt{OneVsRestClassifier}. During prediction, the outputs of the binary classifiers are combined according to the OvR decision rule to obtain the final multiclass prediction.


\section{Experimental Design}

This section describes the experimental design used to evaluate the proposed iCost framework. We first describe the binary and multiclass imbalanced datasets used in the study. Next, we present the validation protocol, preprocessing steps, base classifiers, and compared learning strategies. Finally, we describe the performance metrics used to assess the effectiveness of the proposed approach.

\subsection{Datasets}

The performance of the proposed framework was evaluated on datasets with varying degrees of imbalance to assess its generalizability across different imbalance levels and data distributions. A total of 65 binary and 10 multiclass imbalanced datasets were used in the experiments. The datasets were collected from publicly available benchmark repositories, including the KEEL and UCI data repositories \cite{derrac2015keel}. These datasets cover a wide range of sample sizes, feature dimensions, class distributions, and IRs. The summary of the binary datasets is provided in Table~\ref{tab:binary_datasets}, while the summary of the multiclass datasets is provided in Table~\ref{tab:multiclass_datasets}.



\begin{table}
    \centering
    \caption{Summary of the datasets for binary classification tasks}
    \resizebox{0.55\linewidth}{!}{%
    \begin{tabular}{llll}
    \hline
        \textbf{Dataset Name} & \textbf{\# Samples} & \textbf{\# Features} & \textbf{Imbalance Ratio} \\ \hline
        glass1 & 213 & 10 & 1.8 \\ \hline
        wisconsin & 683 & 10 & 1.86 \\ \hline
        pima & 768 & 9 & 1.87 \\ \hline
        glass0 & 213 & 10 & 2.09 \\ \hline
        yeast1 & 1483 & 9 & 2.46 \\ \hline
        vehicle2 & 846 & 19 & 2.88 \\ \hline
        vehicle1 & 846 & 19 & 2.9 \\ \hline
        vehicle3 & 846 & 19 & 2.99 \\ \hline
        vehicle0 & 845 & 19 & 3.27 \\ \hline
        new-thyroid1 & 215 & 6 & 5.14 \\ \hline
        ecoli2 & 336 & 8 & 5.46 \\ \hline
        glass6 & 214 & 10 & 6.38 \\ \hline
        yeast3 & 1484 & 9 & 8.1 \\ \hline
        yeast & 1484 & 9 & 8.1 \\ \hline
        ecoli3 & 336 & 8 & 8.6 \\ \hline
        page-blocks0 & 5472 & 11 & 8.79 \\ \hline
        ecoli-0-3-4\_vs\_5 & 200 & 8 & 9 \\ \hline
        yeast-2\_vs\_4 & 514 & 9 & 9.08 \\ \hline
        ecoli-0-6-7\_vs\_3-5 & 222 & 8 & 9.09 \\ \hline
        ecoli-0-2-3-4\_vs\_5 & 202 & 8 & 9.1 \\ \hline
        yeast-0-3-5-9\_vs\_7-8 & 506 & 9 & 9.12 \\ \hline
        glass-0-1-5\_vs\_2 & 172 & 10 & 9.12 \\ \hline
        yeast-0-2-5-7-9\_vs\_3-6-8 & 1004 & 9 & 9.14 \\ \hline
        yeast-0-2-5-6\_vs\_3-7-8-9 & 1004 & 9 & 9.14 \\ \hline
        ecoli-0-4-6\_vs\_5 & 203 & 7 & 9.15 \\ \hline
        ecoli-0-2-6-7\_vs\_3-5 & 224 & 8 & 9.18 \\ \hline
        glass-0-4\_vs\_5 & 92 & 10 & 9.22 \\ \hline
        ecoli-0-3-4-6\_vs\_5 & 205 & 8 & 9.25 \\ \hline
        ecoli-0-3-4-7\_vs\_5-6 & 257 & 8 & 9.28 \\ \hline
        vowel & 988 & 14 & 9.98 \\ \hline
        ecoli-0-6-7\_vs\_5 & 220 & 7 & 10 \\ \hline
        glass-0-1-6\_vs\_2 & 192 & 10 & 10.29 \\ \hline
        ecoli-0-1-4-7\_vs\_2-3-5-6 & 336 & 8 & 10.59 \\ \hline
        glass-0-6\_vs\_5 & 108 & 10 & 11 \\ \hline
        glass-0-1-4-6\_vs\_2 & 205 & 10 & 11.06 \\ \hline
        glass2 & 214 & 10 & 11.59 \\ \hline
        ecoli-0-1-4-7\_vs\_5-6 & 332 & 7 & 12.28 \\ \hline
        cleveland-0\_vs\_4 & 177 & 14 & 12.62 \\ \hline
        shuttle-c0-vs-c4 & 1829 & 10 & 13.87 \\ \hline
        yeast-1\_vs\_7 & 459 & 8 & 14.3 \\ \hline
        glass4 & 214 & 10 & 15.46 \\ \hline
        ecoli4 & 336 & 8 & 15.8 \\ \hline
        page-blocks-1-3\_vs\_4 & 472 & 11 & 15.86 \\ \hline
        abalone & 731 & 9 & 16.4 \\ \hline
        glass-0-1-6\_vs\_5 & 184 & 10 & 19.44 \\ \hline
        yeast-1-4-5-8\_vs\_7 & 693 & 9 & 22.1 \\ \hline
        yeast4 & 1484 & 9 & 28.1 \\ \hline
        yeast128 & 947 & 9 & 30.57 \\ \hline
        yeast5 & 1484 & 9 & 32.73 \\ \hline
        winequality-red-8\_vs\_6 & 656 & 12 & 35.44 \\ \hline
        ecoli\_013vs26 & 281 & 8 & 39.14 \\ \hline
        abalone-17\_vs\_7-8-9-10 & 2338 & 9 & 39.31 \\ \hline
        yeast6 & 1483 & 9 & 41.37 \\ \hline
        winequality-white-3\_vs\_7 & 900 & 12 & 44 \\ \hline
        winequality-red-8\_vs\_6-7 & 855 & 12 & 46.5 \\ \hline
        kddcup-land\_vs\_portsweep & 1060 & 39 & 49.48 \\ \hline
        abalone-19\_vs\_10-11-12-13 & 1622 & 9 & 49.69 \\ \hline
        winequality\_white & 1481 & 12 & 58.24 \\ \hline
        poker-8-9\_vs\_6 & 1484 & 11 & 58.36 \\ \hline
        winequality-red-3\_vs\_5 & 691 & 12 & 68.1 \\ \hline
        abalone\_20 & 1916 & 8 & 72.69 \\ \hline
        kddcup-land\_vs\_satan & 1609 & 39 & 79.45 \\ \hline
        poker-8-9\_vs\_5 & 2074 & 11 & 81.96 \\ \hline
        poker\_86 & 1477 & 11 & 85.88 \\ \hline
        kddr\_rookkit & 2225 & 42 & 100.14 \\ \hline
    \end{tabular}
    }%
    \label{tab:binary_datasets}
\end{table}

\begin{table} 
    \centering
     \caption{Summary of the datasets for multiclass classification tasks}
     \resizebox{0.95\linewidth}{!}{%
    \begin{tabular}{llll}
    \hline
         Dataset &  \# Features& \# Classes& No of Samples/Class\\
         \hline
         thyroid &21&  3& 6666, 368, 166\\
         
         new-thyroid &5&  3& 150, 35, 30\\
         
         contraceptive &9&  3& 629, 511, 333\\
         
         shuttle &9&  3& 1706, 338, 131\\
         
         Dry\_Bean\_Dataset &16&  7&  3546, 2636, 2027, 1928, 1630, 1322, 522\\
         
         balance &4&  3& 288, 288, 49\\
         
         pageblocks &10&  5& 4913, 329, 115, 88, 28\\
         
         HCV &12&  4& 540, 30, 24, 21\\
         
         yeast &8&  10& 463, 429, 244, 163, 51, 44, 35, 30, 20, 5\\
         
        wine &13& 3&71, 59, 48\\
 \hline
    \end{tabular}
    }
    \label{tab:multiclass_datasets}
\end{table}

\subsection{Experimental Setup}

To ensure reliable evaluation and avoid data leakage, all preprocessing and model training steps were performed within the cross-validation procedure. A repeated stratified k-fold cross-validation strategy was used, with 5 folds and 10 repetitions, resulting in 50 test evaluations for each dataset. Stratification was applied to preserve the class distribution in each fold. The final performance for each dataset was obtained by averaging the results across all folds and repetitions.

Some datasets contained missing values. These missing entries were imputed before model training to ensure that all classifiers could be applied consistently. Feature scaling was applied using MinMaxScaler. The scaler was fitted only on the training portion of each fold and then applied to the corresponding test portion. This ensured that information from the test fold was not used during preprocessing.

Five base classifiers were used: LR, SVM, DT, RF, and XGBoost. These classifiers were selected to evaluate the effectiveness of iCost across linear, margin-based, tree-based, ensemble-based, and boosting-based learning paradigms. Unless otherwise stated, the default hyperparameter settings of the corresponding ML libraries were used. The base classifier hyperparameters were not tuned, allowing the comparison to focus on the effect of cost-sensitive and instance-complexity-based weighting.

For each classifier, four learning strategies were evaluated: the non-cost-sensitive baseline, the traditional cost-sensitive version, the neighborhood-based iCost variant, and the Gini-impurity-based iCost variant. The default penalty values described in the Methodology section were used throughout the experiments to ensure a consistent and reproducible comparison across datasets.


To provide a broader comparison, several commonly used resampling methods were also included: ROS, RUS, SMOTE, BL-SMOTE, ADASYN, SMOTE-ENN, SMOTE-Tomek, ENN, and Tomek-Links. These methods were implemented using the imbalanced-learn library with default parameter settings. All resampling operations were performed only on the training fold within the cross-validation procedure to avoid data leakage.

\subsection{Performance Metrics}

Assessing the performance of different techniques on skewed data can be challenging \cite{branco2016survey}. ML algorithms often excel at predicting instances from the majority class but tend to perform poorly on the minority class. Consequently, traditional performance metrics like accuracy can be misleading because they do not account for the distribution of classes. 

To illustrate this issue, consider a dataset where 95\% of the samples belong to class-1 and only 5\% belong to class-2. A model that always predicts class-1 will achieve 95\% accuracy but will completely fail to identify class-2. This shows that accuracy can hide poor minority-class performance. In this case, while the specificity will be as high as 100\%, the sensitivity will be 0. Sensitivity and specificity provide class-specific information, representing the performance on the minority and majority classes, respectively. However, each of these metrics describes only one side of the classification performance.

To obtain a broader assessment, composite metrics are commonly used for imbalanced classification \cite{mullick2020appropriateness}. G-mean provides the geometric mean of sensitivity and specificity, while F1-score provides the harmonic mean of precision and sensitivity. These metrics are useful because they combine multiple aspects of classifier behavior. However, they also have limitations. G-mean is based on class-wise rates and may not clearly reflect the absolute number of misclassified samples. F1-score focuses on positive-class prediction performance but does not consider true negatives, which may be important when majority-class misclassifications are also costly. 

To further illustrate this limitation, consider a dataset with 100 minority-class samples and 10,000 majority-class samples. A sensitivity score of 0.8 indicates that the model correctly identifies 80 minority-class samples and misclassifies 20 minority-class samples. Similarly, a specificity score of 0.8 indicates that the model correctly identifies 80\% of the majority-class samples but still misclassifies 2,000 majority-class samples. Thus, the same numerical value of 0.8 corresponds to very different numbers of misclassified instances for the two classes.

This issue also affects composite metrics such as G-mean. Suppose an algorithm improves sensitivity from 0.8 to 0.9, reducing minority-class misclassifications from 20 to 10. However, if this improvement reduces specificity from 0.8 to 0.7, the number of misclassified majority-class samples increases from 2,000 to 3,000. In both cases, the G-mean remains almost the same: \(\sqrt{0.8 \times 0.8}=0.80\) and \(\sqrt{0.9 \times 0.7}\approx0.79\). However, the second model misclassifies nearly 1,000 additional samples overall. This difference is not clearly reflected by G-mean alone.

The same example also illustrates the usefulness of F1-score and MCC. In the first case, where sensitivity and specificity are both 0.8, the model produces TP=80, FN=20, TN=8000, and FP=2000. This results in an F1-score of 0.0734 and an MCC of 0.1469. In the second case, where sensitivity increases to 0.9 but specificity decreases to 0.7, the model produces TP=90, FN=10, TN=7000, and FP=3000. Although the G-mean changes only slightly from 0.8000 to 0.7937, the F1-score drops to 0.0564 and the MCC drops to 0.1289. This shows that the improvement in minority-class sensitivity is offset by a much larger increase in false positives.

To obtain a more reliable assessment, this study considers MCC, F1-score, and ROC-AUC. MCC provides a particularly robust performance measure because it considers all four components of the confusion matrix: true positives (TP), true negatives (TN), false positives (FP), and false negatives (FN) \cite{chicco2020advantages}. A high MCC score is obtained only when the classifier performs well across all cases, making it one of the most balanced measures of classification performance under class imbalance \cite{chicco2021matthews}. Therefore, MCC is used as the primary evaluation criterion in this study. As stated in previous studies \cite{mullick2020appropriateness, newaz2024ibrf}, the performance of algorithms on imbalanced data cannot be sufficiently represented by a single metric. Therefore, eight performance metrics are reported in this study: MCC, ROC-AUC, G-mean, F1-score, sensitivity, specificity, precision, and accuracy.

For multiclass settings, three averaging criteria are commonly used: macro, micro, and weighted averaging \cite{grandini2020metrics}. Macro averaging gives equal importance to each class and is therefore suitable for imbalanced datasets, since poor performance on less frequent classes is not overshadowed by strong performance on majority classes. Therefore, macro-averaged results are calculated and reported for the multiclass experiments.

The  metrics used in this study are defined as follows:

\begin{equation}
Sensitivity = \frac{TP}{TP+FN},
\end{equation}

\begin{equation}
Specificity = \frac{TN}{TN+FP},
\end{equation}

\begin{equation}
Precision = \frac{TP}{TP+FP},
\end{equation}

\begin{equation}
Accuracy = \frac{TP+TN}{TP+TN+FP+FN},
\end{equation}

\begin{equation}
F1\text{-}score = \frac{2 \times Precision \times Sensitivity}{Precision + Sensitivity},
\end{equation}

\begin{equation}
G\text{-}mean = \sqrt{Sensitivity \times Specificity},
\end{equation}

\begin{equation}
MCC = \frac{TP \times TN - FP \times FN}
{\sqrt{(TP+FP)(TP+FN)(TN+FP)(TN+FN)}}.
\end{equation}

\section{Results and Discussion}

In this section, we present and discuss the results obtained from our experiments. Since we are working with a large number of datasets, different performance measures on individual datasets cannot be shown here in the manuscript; they are provided in the Supplementary Files S1--S4. 


\subsection{Overall Performance on Binary Datasets}

Standard classification algorithms often perform poorly on imbalanced datasets. The performance gets even worse with higher imbalances and class overlapping \cite{mohosheu2024comprehensive}. In 19 out of 65 datasets, the G-mean score was found to be 0 for the SVM classifier. This indicates that the classifier failed to identify any minority-class samples in those cases, producing a sensitivity of 0 while maintaining perfect or near-perfect specificity. Such results reflect a clear bias toward the majority class.

Traditional CSL helps reduce this bias by assigning a higher penalty to minority-class misclassifications. However, this improvement comes with a trade-off. While CSL generally improves sensitivity, it reduces specificity, sometimes substantially, by increasing the number of majority-class samples misclassified as minority-class samples. This happens because conventional CSL assigns the same penalty to all minority-class instances, regardless of whether they are boundary or noisy samples. Consequently, although CSL can improve minority-class recognition, it also introduces unnecessary decision-boundary distortion. 


The proposed iCost framework aims to mitigate this limitation by assigning penalties more carefully at the instance level. Instead of uniformly penalizing all minority-class samples, iCost adapts the penalty according to the estimated complexity of each instance. We start with the default cost settings and observe its performance.  Tables~\ref{tab:binary_mcc} and~\ref{tab:binary_f1} present the average MCC and F1-score obtained across the 65 binary imbalanced datasets, respectively. As shown in Table~\ref{tab:binary_mcc}, both proposed variants outperform traditional CSL for all five classifiers in terms of average MCC. The overall average MCC improves from 53.89 with traditional CSL to 55.36 with Neighbor-iCost and 54.49 with Gini-iCost. Similarly, the average F1-score improves from 56.01 with traditional CSL to 57.74 and 56.77 for Neighbor-iCost and Gini-iCost, respectively. These results indicate that instance-complexity-aware penalization provides a more balanced learning strategy than uniform class-level penalization.


It should also be noted that for RF, the non-cost-sensitive baseline achieves a slightly higher average MCC than the cost-sensitive variants, suggesting that RF is less dependent on explicit cost weighting. However, Neighbor-iCost still improves over traditional CSL for RF, indicating that instance-level penalization is more effective than uniform class-level penalization.

\begin{table}[!htbp]
\centering
\caption{Average MCC scores across 65 binary imbalanced datasets (in percentage).}
\label{tab:binary_mcc}
\begin{tabular}{lcccc}
\hline
\textbf{Classifier} & \textbf{NCS} & \textbf{CS} & \textbf{Neighbor-iCost} & \textbf{Gini-iCost} \\
\hline
LR       & 26.63 & 48.22 & \textbf{51.07} & 48.88 \\
SVM      & 46.23 & 58.15 & \textbf{60.06} & 59.36 \\
DT       & 51.61 & 50.98 & \textbf{52.30} & 51.38 \\
RF       & \textbf{54.56} & 53.80 & 54.19 & 53.96 \\
XGBoost  & 55.02 & 58.28 & \textbf{59.20} & 58.85 \\
\hline
\textbf{Average} & 46.81 & 53.89 & \textbf{55.36} & 54.49 \\
\hline
\end{tabular}
\end{table}

\begin{table}[!htbp]
\centering
\caption{Average F1-scores across 65 binary imbalanced datasets (in percentage).}
\label{tab:binary_f1}
\begin{tabular}{lcccc}
\hline
\textbf{Classifier} & \textbf{NCS} & \textbf{CS} & \textbf{Neighbor-iCost} & \textbf{Gini-iCost} \\
\hline
LR       & 25.97 & 50.60 & \textbf{54.33} & 51.71 \\
SVM      & 46.80 & 60.16 & \textbf{62.60} & 61.70 \\
DT       & 55.06 & 54.22 & \textbf{55.64} & 54.67 \\
RF       & \textbf{55.19} & 54.19 & 54.52 & 54.37 \\
XGBoost  & 56.25 & 60.88 & \textbf{61.61} & 61.39 \\
\hline
\textbf{Average} & 47.86 & 56.01 & \textbf{57.74} & 56.77 \\
\hline
\end{tabular}
\end{table}

\subsection{Win-Loss Analysis Against Traditional CSL}

Although average performance provides an overall view of the effectiveness of the proposed framework, it does not fully show how consistently the method improves across individual datasets. Therefore, a win-loss analysis was conducted by comparing each proposed iCost variant with traditional CSL on each of the 65 binary imbalanced datasets. A win indicates that the proposed method achieved a higher MCC than traditional CSL for a given dataset and classifier.

\begin{table}[!htbp]
\centering
\caption{Win-loss analysis of the proposed iCost variants against traditional CSL based on MCC across 65 binary imbalanced datasets.}
\label{tab:win_loss_cs}
\begin{tabular}{llcccc}
\hline
\textbf{Method} & \textbf{Classifier} & \textbf{iCost Wins} & \textbf{CS Wins} & \textbf{Ties} \\
\hline
Neighbor-iCost & LR       & 46 & 15 & 4   \\
Neighbor-iCost & SVM      & 39 & 16 & 10  \\
Neighbor-iCost & DT       & 31 & 28 & 6   \\
Neighbor-iCost & RF       & 33 & 24 & 8  \\
Neighbor-iCost & XGBoost  & 34 & 26 & 5  \\
\hline
Gini-iCost     & LR       & 40 & 22 & 3   \\
Gini-iCost     & SVM      & 36 & 17 & 12  \\
Gini-iCost     & DT       & 26 & 32 & 7   \\
Gini-iCost     & RF       & 23 & 23 & 19  \\
Gini-iCost     & XGBoost  & 34 & 27 & 4  \\
\hline
\end{tabular}
\end{table}

Table~\ref{tab:win_loss_cs} shows that Neighborhood-based iCost outperforms traditional CSL more frequently than it loses. The improvement is particularly strong for LR and SVM, where Neighbor-iCost wins on 46 and 39 datasets, respectively. This suggests that adaptive instance-level penalization is especially beneficial for classifiers that are more sensitive to the weighting of training samples. Tree-based ensembles are more robust and less sensitive to instance weighting. However, on average, Neighbor-iCost remains quite competitive for ensemble-based classifiers. Gini-impurity-based feature space partitioning also shows competitive performance against traditional CSL. It wins in the majority of the datasets for LR, SVM, and XGBoost, while its performance is more balanced for DT and RF. This indicates that the Gini-impurity-based strategy can provide useful regional information about instance complexity, although its improvement is less consistent than the neighborhood-based strategy.

Overall, across all 325 classifier-dataset combinations, Neighbor-iCost wins in 183 cases, loses in 109 cases, and ties in 33 cases against traditional CSL. Gini-iCost wins in 159 cases, loses in 121 cases, and ties in 45 cases. These results confirm that the proposed instance-complexity-based penalty assignment is not only beneficial on average, but also improves performance consistently across a wide range of imbalanced datasets. Furthermore, these results are obtained using the initial default cost settings without any tuning. How performance varies with different penalty values is explored in the subsequent sections.

\subsection{Trade-off Between Sensitivity and Specificity}

Traditional CSL improves minority-class recognition by assigning a higher penalty to minority-class misclassifications. However, this often increases sensitivity at the cost of specificity. In highly imbalanced datasets, even a small reduction in specificity can produce a large number of FPs because the majority class contains substantially more samples than the minority class. Although correctly identifying minority-class instances (positive cases) is often important, a prediction framework may lose reliability if this improvement causes a large increase in majority-class misclassifications. Therefore, improving sensitivity alone is not always sufficient; a useful imbalanced learning method should improve minority-class recognition without causing excessive majority-class misclassifications.

To examine this trade-off more closely, we counted the confusion matrix parameters (TP, TN, FP, and FN), for all datasets. Table~\ref{tab:sens_spec_tradeoff} summarizes the total changes obtained by Neighbor-iCost and Gini-iCost compared to traditional CSL approach across all datasets. The confusion matrix parameters for individual datasets are provided as a Supplementary File -- S1.

\begin{table}[!ht]
\centering
\caption{Error trade-off of Neighbor- and Gini-based iCost approach compared with traditional CSL across 65 binary datasets. Negative values indicate reductions}
\label{tab:sens_spec_tradeoff}
\begin{tabular}{lrrrrrr}
\hline
\textbf{Classifier}
& \multicolumn{3}{c}{\textbf{Neighbor-iCost vs CS}}
& \multicolumn{3}{c}{\textbf{Gini-iCost vs CS}} \\
\cline{2-7}
& \textbf{$\Delta$FP} & \textbf{$\Delta$FN} & \textbf{$\Delta$Correct}
& \textbf{$\Delta$FP} & \textbf{$\Delta$FN} & \textbf{$\Delta$Correct} \\
\hline
LR       & -3476 &  282 & 3194 & -3394 & 278 & 3116 \\
SVM      & -1467 &  131 & 1336 & -1178 & 129 & 1048 \\
DT       &    49 &  -22 &  -26 &    29 &   6 &  -36 \\
RF       &     5 &  -18 &   13 &    22 & -24 &    2 \\
XGBoost  &  -412 &   78 &  333 &  -256 &  46 &  210 \\
\hline
\end{tabular}
\end{table}

The results show that both Neighbor-iCost and Gini-iCost substantially reduce FPs for LR, SVM, and XGBoost classifiers. For example, in the SVM setting, Neighbor-iCost increases the overall number of FNs by approximately 131 but reduces FPs by 1467. The reduction in FPs is more than eleven times larger, leading to approximately 1336 fewer total incorrect classifications. A similar pattern is observed for LR, where FPs decrease by 3476 while FNs increase by 282. These results suggest that Neighbor-iCost reduces the excessive majority-class misclassifications caused by uniform minority-class penalization. Gini-iCost shows a similar but slightly weaker trend. It reduces FPs by 3394, 1178, and 256 for LR, SVM, and XGBoost, respectively. However, for RF, the trade-off is less pronounced, indicating that such a robust ensemble method benefits less from additional instance-level cost adjustment.

Unlike traditional CSL, which may over-expand the minority decision region by uniformly penalizing all minority-class samples, our proposed approach assigns lower penalties to easy or noisy samples and higher penalties to informative boundary samples. This reduces unnecessary majority-class misclassifications while preserving useful minority-class emphasis. As a result, iCost slightly increases FNs, but this increase is relatively small compared with the substantial reduction in FPs. Therefore, the proposed method does not excessively sacrifice minority-class recognition; rather, it reduces unnecessary majority-class misclassifications while maintaining almost similar emphasis on difficult minority-class samples.


\subsection{Stability and Computational Cost}

Table~\ref{tab:stability_runtime_classifier} summarizes the classifier-wise stability and computational cost of the cost-sensitive approaches. The MCC standard deviation was calculated across the repeated stratified k-fold cross-validation runs, providing an indication of how stable each method is across different train-test splits. The average MCC standard deviations of the proposed methods are slightly higher than that of traditional CSL, which is expected because instance-level weighting depends on the local or regional structure of each training fold. However, the increase is modest, and the proposed methods maintain higher average MCC performance.

In terms of computational cost, Neighbor-iCost generally introduces additional overhead because it requires nearest-neighbor analysis before assigning instance-level weights. Gini-iCost has a smaller overhead, since it only requires fitting a shallow decision-tree probe and computing leaf-level impurity. Nevertheless, the fit times remain small for all cost-sensitive approaches. Even for RF, which has the largest training time among the tested classifiers, the average fit time remains below one second. This indicates that the additional instance-complexity estimation step does not create a substantial computational burden. From a practical perspective, the small training time also suggests that light cost-parameter tuning is feasible, allowing iCost to be used both with fixed default costs and with tuned cost settings when further performance improvement is desired.


\begin{table}[!htbp]
\centering
\small
\caption{Stability and computational cost across 65 binary datasets.}
\label{tab:stability_runtime_classifier}
\begin{tabular}{lrrrrrr}
\hline
\textbf{Classifier}
& \multicolumn{3}{c}{\textbf{MCC Std.}}
& \multicolumn{3}{c}{\textbf{Fit Time (s)}} \\
\cline{2-7}
& \textbf{CS} & \textbf{Neighbor} & \textbf{Gini}
& \textbf{CS} & \textbf{Neighbor} & \textbf{Gini} \\
\hline
LR      & 0.0915 & 0.1109 & 0.1104 & 0.0079 & 0.0167 & 0.0110 \\
SVM     & 0.1026 & 0.1147 & 0.1107 & 0.0210 & 0.0247 & 0.0221 \\
DT      & 0.1379 & 0.1432 & 0.1375 & 0.0073 & 0.0116 & 0.0106 \\
RF      & 0.1216 & 0.1237 & 0.1267 & 0.2446 & 0.3003 & 0.2530 \\
XGBoost & 0.1360 & 0.1408 & 0.1398 & 0.0419 & 0.0635 & 0.0465 \\
\hline
\textbf{Average} & 0.1179 & 0.1267 & 0.1250 & 0.0645 & 0.0834 & 0.0686 \\
\hline
\end{tabular}
\end{table}

\subsection{Cost-Parameter Sensitivity Analysis}

\subsubsection{Tuning Setup}
The main experiments used a fixed, conceptually motivated cost hierarchy to ensure a consistent evaluation protocol across datasets and classifiers. In the default setting, outlier-like minority instances receive the lowest penalty, pure/easy instances receive a moderate penalty, safe instances receive a higher penalty, and borderline instances receive the highest penalty. This reflects the central assumption of iCost: minority samples should not be penalized uniformly because their contribution to learning the decision boundary differs according to their estimated complexity.

To examine the effect of cost selection, a cost-parameter sensitivity analysis was conducted. This is important because the default iCost configuration was designed based on the intuitive complexity hierarchy. However, the exact penalty magnitudes may influence model behavior differently across classifiers and datasets. Therefore, a grid-search analysis was used to observe how different penalty values are associated with the best predictive performance. The four instance-level cost parameters were varied independently using three candidate values for each parameter. Table~\ref{tab:cost_grid} shows the cost-parameter search space used in the tuning experiment. This resulted in \(3^4=81\) cost combinations for each iCost variant. The best setting for each dataset was selected based on MCC, with F1-score and G-mean used as tie-breakers. To reduce computational burden, the grid-search analysis was performed using a single 5-fold stratified cross-validation. To provide a fair comparison, the global minority-class penalty of traditional CSL was also tuned using three penalty values, as reported in the table.

\begin{table}[!htbp]
\centering
\caption{Cost-parameter search space used in the tuning experiment.}
\label{tab:cost_grid}
\begin{tabular}{lll}
\hline
\textbf{Method} & \textbf{Parameter} & \textbf{Candidate values x IR} \\
\hline
Traditional CSL & Minority penalty & ${0.8, 1.0, 1.2}$ \\
\hline
iCost & $cfo (Outlier)$ & ${0.10, 0.20, 0.30}$ \\
& $cfp (Pure)$ & ${0.30, 0.50, 0.70}$ \\
& $cfs (Safe)$ & ${0.75, 0.90, 1.10}$ \\
& $cfb (Border)$ & ${1.00, 1.25, 1.50}$ \\
\hline
\end{tabular}
\end{table}

\subsubsection{Effect of Cost Tuning}
The cost-parameter tuning results are reported in Table~\ref{tab:tuning_mcc_results}. In the previous fixed-cost experiments, the proposed iCost variants produced moderate improvements over traditional CSL, showing that instance-level penalty assignment is useful even with a simple default cost hierarchy. However, the tuned results show that the performance gain can be substantially larger when the penalty values are adapted appropriately through grid-search. This indicates that the original cost setting provides a reasonable baseline, but it is not necessarily the optimal configuration for every classifier and dataset. 

For traditional CSL, tuning the global minority-class penalty improves MCC for all classifiers. The improvement ranges from +0.78 for RF to +2.44 for LR, suggesting that 
a single global IR-based penalty is not always optimal. However, the gains from tuning traditional CSL remain relatively limited because the same penalty is assigned to all minority samples. In contrast, the proposed iCost variants usually obtain larger gains after tuning. Neighbor-iCost improves MCC by +3.03, +1.92, +5.51, +4.36, and +3.80 for LR, SVM, DT, RF, and XGBoost, respectively. Similarly, Gini-iCost improves MCC by +2.90, +1.77, +5.40, +2.98, and +4.71 for LR, SVM, DT, RF, and XGBoost, respectively.



\begin{table}[!htbp]
\centering
\small
\caption{MCC scores (in percentage) obtained after cost-parameter tuning.}
\label{tab:tuning_mcc_results}
\begin{tabular}{llccc}
\hline
\textbf{Classifier} & \textbf{Method} & \textbf{Default} & \textbf{Tuned} & \textbf{$\Delta$MCC} \\
\hline
LR & CSL & 48.11 & 50.55 & 2.44 \\
LR & Neighbor-iCost & 51.90 & 54.93 & 3.03 \\
LR & Gini-iCost & 49.56 & 52.47 & 2.90 \\
\hline
SVM & CSL & 58.70 & 59.81 & 1.10 \\
SVM & Neighbor-iCost & 60.26 & 62.18 & 1.92 \\
SVM & Gini-iCost & 60.61 & 62.39 & 1.77 \\
\hline
DT & CSL & 51.14 & 53.47 & 2.33 \\
DT & Neighbor-iCost & 52.09 & 57.60 & 5.51 \\
DT & Gini-iCost & 51.38 & 56.78 & 5.40 \\
\hline
RF & CSL & 53.83 & 54.61 & 0.78 \\
RF & Neighbor-iCost & 53.04 & 57.41 & 4.36 \\
RF & Gini-iCost & 54.42 & 57.40 & 2.98 \\
\hline
XGBoost & CSL & 58.15 & 59.67 & 1.52 \\
XGBoost & Neighbor-iCost & 59.73 & 63.53 & 3.80 \\
XGBoost & Gini-iCost & 58.82 & 63.54 & 4.71 \\
\hline
\end{tabular}
\end{table}

\subsubsection{Tuned iCost versus Tuned CSL}
\begin{figure}[!htbp]
\centering
\includegraphics[width=0.85\textwidth]{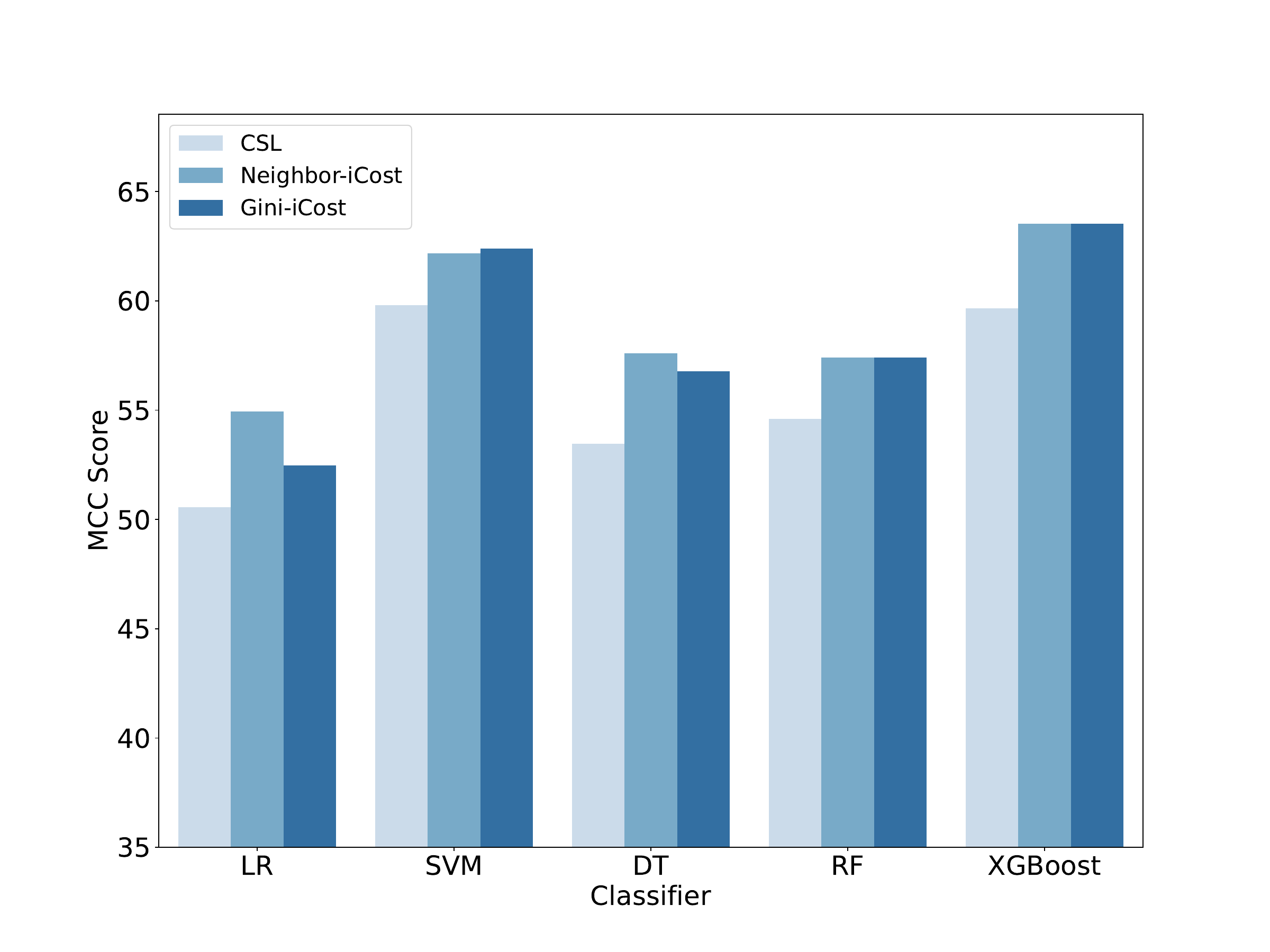}
\caption{Average MCC score (in percentage) comparison of tuned CSL, Neighbor-iCost, and Gini-iCost across 65 binary imbalanced datasets.}
\label{fig:tuned_icost_mcc_bar}
\end{figure}

Table~\ref{tab:tuned_icost_vs_tuned_csl_winloss} compares the tuned iCost variants with tuned traditional CSL. In the previous fixed-cost experiments, iCost produced moderate but consistent improvements over traditional CSL, although the win-loss behavior was less strong for some classifiers (Table \ref{tab:win_loss_cs}). For example, the default Neighbor-iCost setting showed clear improvements for LR and SVM, but the margins were smaller for DT, RF, and XGBoost. Similarly, default Gini-iCost showed mixed behavior for some tree-based classifiers. After cost tuning, however, the advantage of iCost becomes substantially clearer. Figure~\ref{fig:tuned_icost_mcc_bar} visually summarizes these average MCC improvements across classifiers.

For Neighbor-iCost, the average MCC improvement over tuned CSL ranges from +2.38 to +4.37 points across classifiers. It wins against tuned CSL on 54 datasets for LR, 40 for SVM, 43 for DT, 45 for RF, and 51 for XGBoost. This is a notable improvement over the fixed-cost setting. Gini-iCost also improves over tuned CSL. This is especially important because default Gini-iCost had only modest gains in some earlier comparisons. After tuning, Gini-iCost wins the majority of datasets, including 50 wins against only 8 losses for XGBoost. This indicates that the Gini-impurity-based regional complexity estimate is useful, but the associated cost values need to be calibrated appropriately. Across all 325 classifier-dataset combinations, tuned Neighbor-iCost wins in 233 cases, loses in 45 cases, and ties in 47 cases against tuned CSL. Tuned Gini-iCost wins in 212 cases, loses in 66 cases, and ties in 47 cases.




\begin{table}[!htbp]
\centering
\small
\caption{Win-loss analysis of tuned iCost variants against tuned traditional CSL based on MCC across 65 binary imbalanced datasets.}
\label{tab:tuned_icost_vs_tuned_csl_winloss}
\begin{tabular}{llcccc}
\hline
\textbf{Method} & \textbf{Classifier} & \textbf{Avg. \(\Delta\)MCC} & \textbf{iCost Wins} & \textbf{CS Wins} & \textbf{Ties} \\
\hline
Neighbor-iCost & LR      & 4.37 & 54 & 7  & 4  \\
Neighbor-iCost & SVM     & 2.38 & 40 & 10 & 15 \\
Neighbor-iCost & DT      & 4.13 & 43 & 16 & 6  \\
Neighbor-iCost & RF      & 2.80 & 45 & 6  & 14 \\
Neighbor-iCost & XGBoost & 3.86 & 51 & 6  & 8  \\
\hline
Gini-iCost     & LR      & 1.92 & 42 & 20 & 3  \\
Gini-iCost     & SVM     & 2.58 & 41 & 13 & 11 \\
Gini-iCost     & DT      & 3.31 & 39 & 20 & 6  \\
Gini-iCost     & RF      & 2.79 & 40 & 5 & 20 \\
Gini-iCost     & XGBoost & 3.87 & 50 & 8  & 7  \\
\hline
\end{tabular}
\end{table}

\begin{figure*} 
\centering
\includegraphics[width=\textwidth]{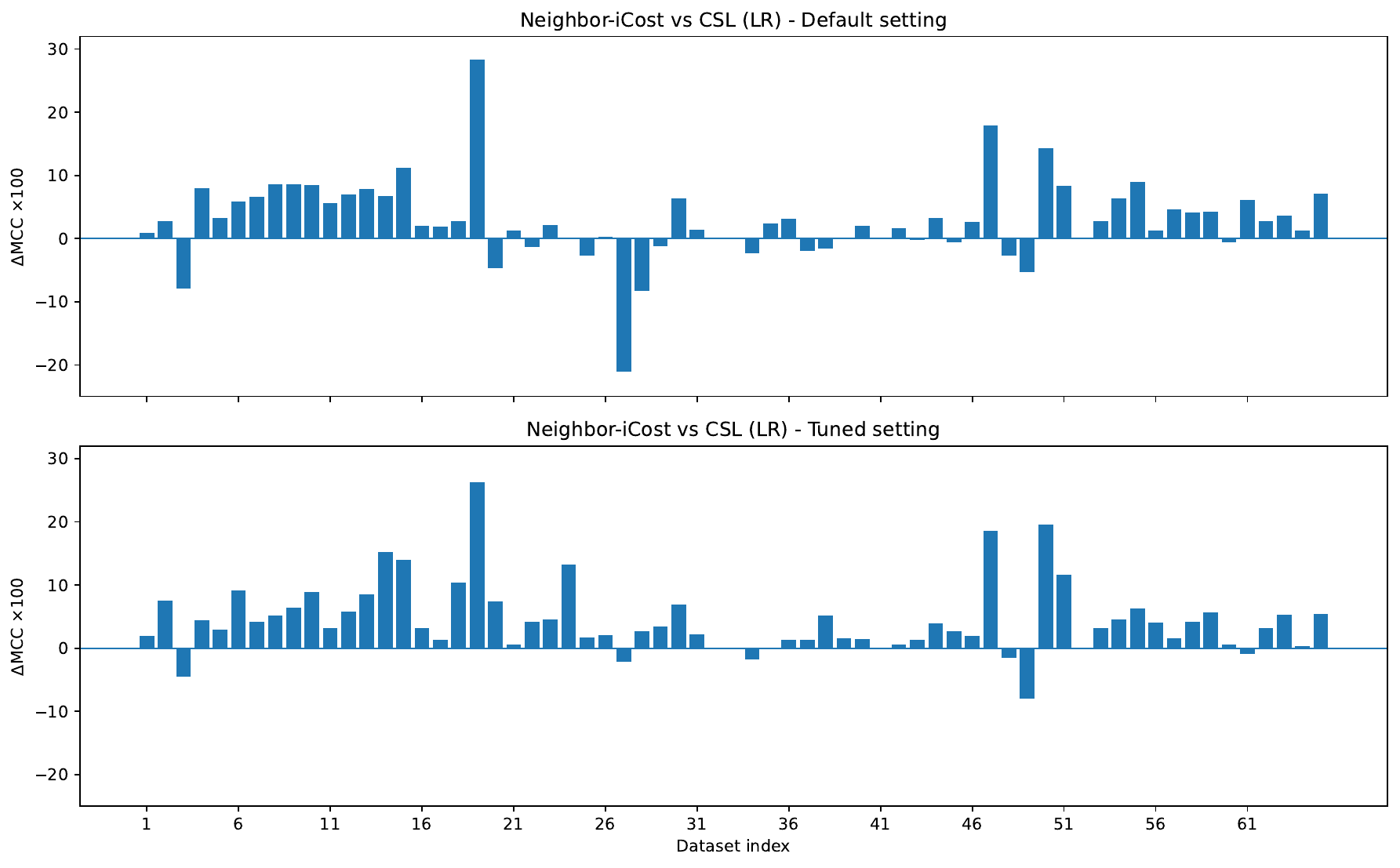}
\caption{Dataset-wise change in MCC of Neighbor-iCost over CSL for the LR classifier under default and tuned cost settings. Positive values indicate that Neighbor-iCost outperforms CSL.}
\label{fig: icost_lr_neighbor}
\end{figure*}

Figure~\ref{fig: icost_lr_neighbor} illustrates the dataset-wise change in MCC obtained by Neighbor-iCost over CSL for the LR classifier under default and tuned cost settings. The default setting already produces positive improvements for most datasets. In many datasets, substantial gains in the MCC scores are observed (more than 5 MCC points). For a few datasets, there was no change in performance. After tuning, the improvement pattern becomes more consistent: several negative or weak-gain cases are reduced, while many positive gains are maintained or further increased. This indicates that cost tuning can strengthen the proposed framework by better adapting the relative penalties of different minority-instance categories to the data distribution. This also confirms that the average improvement reported in Table~\ref{tab:binary_mcc} and Table \ref{tab:tuning_mcc_results} is not driven by a few isolated cases, but reflects consistent improvements across many imbalanced datasets.

\subsubsection{Analysis of Selected Penalty Values}

Tables~\ref{tab:neighbor_cost_frequency} and~\ref{tab:gini_cost_frequency} summarize how often each individual cost value was selected in the best-performing tuned iCost configuration. A clear pattern emerges across both Neighbor-iCost and Gini-iCost: lower penalty values are selected more frequently for most cost components. For cfo, the lowest candidate value 0.10IR is selected most often across all classifiers and both iCost variants. This supports the original motivation that outlier-like minority samples should receive a relatively small penalty.

A similar trend is observed for cfp and cfs. For cfp, the lowest value 0.30IR is selected most frequently in nearly all cases, especially for Neighbor-iCost, where it is selected 60 times for SVM. This suggests that pure minority samples do not require strong penalization, since they are already located in relatively reliable minority regions. For cfs, the lowest value 0.75IR is also selected most often for both iCost variants, indicating that safe minority samples benefit from a moderate penalty, but not necessarily from stronger cost amplification.

For cfb, the lowest candidate value 1.00IR is usually selected most frequently, although the distribution is more spread out than for cfp and cfs. This is expected because borderline samples are more sensitive to classifier behavior and dataset geometry. In some cases, such as DT and XGBoost, higher cfb values are selected more often than in LR or SVM, suggesting that the optimal borderline penalty can be more data- and classifier-dependent. Overall, these results indicate that the default hierarchy of iCost is reasonable, but the exact penalty magnitudes often benefit from a milder configuration. 


\begin{table}[!htbp]
\centering
\small
\setlength{\tabcolsep}{4pt}
\caption{Individual cost-value selection frequency for tuned Neighbor-iCost for 65 datasets. Each value indicates the number of datasets for which the corresponding penalty value was selected in the best-performing cost setting. All candidate values are expressed as multipliers of IR.}
\label{tab:neighbor_cost_frequency}
\begin{tabular}{l|ccc|ccc|ccc|ccc}
\hline
\textbf{Classifier}
& \multicolumn{3}{c|}{\textbf{cfo}}
& \multicolumn{3}{c|}{\textbf{cfp}}
& \multicolumn{3}{c|}{\textbf{cfs}}
& \multicolumn{3}{c}{\textbf{cfb}} \\
\cline{2-13}
& 0.10 & 0.20 & 0.30
& 0.30 & 0.50 & 0.70
& 0.75 & 0.90 & 1.10
& 1.00 & 1.25 & 1.50 \\
\hline
LR      & 47 & 12 & 6  & 53 & 9  & 3  & 52 & 6  & 7  & 40 & 14 & 11 \\
SVM     & 39 & 16 & 10 & 60 & 2  & 3  & 59 & 6  & 0  & 57 & 7  & 1  \\
DT      & 32 & 12 & 21 & 54 & 7  & 4  & 41 & 12 & 12 & 33 & 12 & 20 \\
RF      & 50 & 7  & 8  & 51 & 5  & 9  & 41 & 14 & 10 & 43 & 9  & 13 \\
XGBoost & 40 & 15 & 10 & 47 & 9  & 9  & 36 & 18 & 11 & 35 & 18 & 12 \\
\hline
\end{tabular}
\end{table}

\begin{table}[!htbp]
\centering
\small
\setlength{\tabcolsep}{4pt}
\caption{Individual cost-value selection frequency for tuned Gini-iCost for 65 datasets. Each value indicates the number of datasets for which the corresponding penalty value was selected in the best-performing cost setting. All candidate values are expressed as multipliers of IR.}
\label{tab:gini_cost_frequency}
\begin{tabular}{l|ccc|ccc|ccc|ccc}
\hline
\textbf{Classifier}
& \multicolumn{3}{c|}{\textbf{cfo}}
& \multicolumn{3}{c|}{\textbf{cfp}}
& \multicolumn{3}{c|}{\textbf{cfs}}
& \multicolumn{3}{c}{\textbf{cfb}} \\
\cline{2-13}
& 0.10 & 0.20 & 0.30
& 0.30 & 0.50 & 0.70
& 0.75 & 0.90 & 1.10
& 1.00 & 1.25 & 1.50 \\
\hline
LR      & 43 & 12 & 10 & 44 & 9  & 12 & 53 & 5  & 7  & 42 & 11 & 12 \\
SVM     & 44 & 10 & 11 & 44 & 11 & 10 & 57 & 5  & 3  & 50 & 11 & 4  \\
DT      & 34 & 10 & 21 & 35 & 14 & 16 & 37 & 14 & 14 & 40 & 13 & 12 \\
RF      & 46 & 11 & 8  & 44 & 12 & 9  & 47 & 12 & 6  & 44 & 9  & 12 \\
XGBoost & 41 & 13 & 11 & 35 & 15 & 15 & 34 & 21 & 10 & 35 & 14 & 16 \\
\hline
\end{tabular}
\end{table}

It should be noted that the present tuning analysis used a compact search space, with only three candidate values for each cost parameter. A finer search could potentially identify better cost combinations and further improve performance. However, the objective of this study was not to exhaustively optimize the penalty values, but to examine whether the proposed complexity-aware cost assignment remains beneficial under reasonable cost variations and to identify suitable cost settings.

These findings also highlight a key limitation of traditional CSL. In standard CSL, all minority samples are usually assigned the same global penalty based on the IR. The cost-selection patterns in Tables~\ref{tab:neighbor_cost_frequency} and~\ref{tab:gini_cost_frequency} suggest that this uniform assumption is flawed. The best-performing iCost configurations frequently assign substantially different penalties to different minority-instance categories, and these differentiated penalties lead to better MCC performance than tuned CSL. This supports the main premise of iCost: minority samples should not be treated as equally difficult simply because they belong to the same class; rather, their contribution to learning should be adjusted according to their estimated complexity.

\subsection{Comparison with Resampling Methods}

Resampling is a widely used strategy for handling class imbalance. Unlike CSL, which modifies the learning objective through penalty weights, resampling methods modify the training distribution by either adding minority samples, removing majority samples, or combining both operations. Therefore, comparing iCost with resampling-based pipelines provides an additional perspective on whether complexity-aware instance weighting can serve as an effective alternative to data-level imbalance handling.

In this experiment, the proposed iCost variants were compared with several commonly used resampling methods. All resampling operations were applied only to the training fold within the cross-validation pipeline to avoid data leakage. The performance measures are reported in Table~\ref{tab:resampling_avg_mcc}. Among the resampling methods, SMOTE and SMOTE-Tomek provide the strongest overall performance, with average MCC scores of 57.27 and 57.23, respectively. This indicates that synthetic minority oversampling, either alone or combined with Tomek link cleaning, is a strong baseline for imbalanced classification. 

The iCost variants remain highly competitive against these resampling pipelines. Neighbor-iCost obtains the highest overall average MCC of 59.13, while tuned Gini-iCost achieves 58.51. Both outperform the best average resampling baseline. Classifier-wise, Neighbor-iCost gives the best MCC for LR and DT, while Gini-iCost gives the best MCC for SVM and XGBoost. For XGBoost, the best resampling MCC was 60.44 with SMOTE, whereas both iCost variants achieved MCC values of approximately 63.5. These results suggest that complexity-aware cost assignment can provide performance gains that are comparable to, and in several cases stronger than, those obtained by modifying the training distribution through resampling. 

The main exception is RF, where SMOTE achieves the best MCC of 60.03, while iCost variants obtain around 57.40. This suggests that RF benefits more from synthetic minority expansion. Nevertheless, iCost shows stronger overall average performance across classifiers and has the advantage of preserving the original training distribution. Unlike resampling methods, iCost does not generate synthetic samples or remove existing samples; instead, it adjusts the learning emphasis of minority instances according to their estimated complexity. This is practically important because resampling can become computationally expensive for large datasets, especially when synthetic samples are generated or when undersampling/cleaning methods are repeatedly applied. In contrast, once the cost parameters are selected, iCost can be applied directly through instance-level sample weights without changing the size or composition of the training set. Therefore, the results indicate that iCost can serve as an effective alternative to resampling, particularly when the dataset is large, computational efficiency is important, or modifying the original training distribution is undesirable.

\begin{table}[!htbp]
\centering
\small
\setlength{\tabcolsep}{5pt}
\caption{Average MCC score comparison between resampling methods and iCost variants across 65 binary imbalanced datasets.}
\label{tab:resampling_avg_mcc}
\begin{tabular}{lcccccc}
\hline
\textbf{Method} & \textbf{LR} & \textbf{SVM} & \textbf{DT} & \textbf{RF} & \textbf{XGBoost} & \textbf{Average} \\
\hline
ROS & 52.19 & 59.39 & 51.06 & 58.12 & 57.97 & 55.75 \\
RUS & 46.94 & 49.50 & 41.46 & 49.85 & 45.05 & 46.56 \\
SMOTE & 53.41 & 60.24 & 52.25 & \textbf{60.03} & 60.44 & 57.27 \\
BL-SMOTE & 53.44 & 59.49 & 52.20 & 58.26 & 59.93 & 56.66 \\
ADASYN & 50.63 & 57.63 & 52.88 & 59.25 & 59.27 & 55.93 \\
SMOTE-ENN & 52.11 & 58.09 & 52.20 & 59.09 & 58.29 & 55.96 \\
SMOTE-Tomek & 53.52 & 60.24 & 52.34 & 59.81 & 60.25 & 57.23 \\
ENN & 51.26 & 52.36 & 52.02 & 56.53 & 55.77 & 53.59 \\
Tomek-Links & 50.04 & 50.17 & 51.98 & 55.44 & 55.94 & 52.72 \\
\hline
Neighbor-iCost & \textbf{54.93} & 62.18 & \textbf{57.60} & 57.41 & 63.53 & \textbf{59.13} \\
Gini-iCost & 52.47 & \textbf{62.39} & 56.78 & 57.40 & \textbf{63.54} & 58.51 \\
\hline
\end{tabular}
\end{table}

\subsection{Multiclass Classification Results}

The proposed algorithm can also be directly utilized in multiclass imbalanced scenarios. A total of 10 datasets with different numbers of classes were utilized in the experiment. For multiclass learning, each problem was decomposed into multiple one-vs-rest (OvR) binary subproblems, and the corresponding instance-level weights were computed separately for each binary subproblem. 

Table~\ref{tab:multiclass_mcc} presents the average MCC scores obtained across the datasets. Detailed performance measures on individual datasets are provided in the supplementary File -- S4. Compared with NCS, all cost-sensitive approaches improve the average MCC, indicating that CSL remains useful in multiclass imbalanced settings. Traditional CSL improves the average MCC score from 70.94 to 73.47. Neighbor-iCost achieves a higher average MCC of 73.88, while Gini-iCost obtains an average MCC of 73.71. The classifier-wise results show that Gini-iCost achieves the highest MCC for all classifiers except LR, whereas neighbor-iCost performs the best for LR. However, the differences in performance between the Neighbor and Gini-based approaches are relatively small. The results suggest that the proposed framework remains applicable in multiclass settings, although the performance gains are less pronounced than in binary classification.

\begin{table}[!htbp]
\centering
\small
\caption{Average MCC scores (in percentage) across 10 multiclass imbalanced datasets}
\label{tab:multiclass_mcc}
\begin{tabular}{lcccc}
\hline
\textbf{Classifier} & \textbf{No-cost} & \textbf{Tuned CSL} & \textbf{Neighbor-iCost} & \textbf{Gini-iCost} \\
\hline
LR & 56.63 & 65.31 & \textbf{65.42} & 64.21 \\ 
SVM & 70.69 & 72.96 & 72.87 & \textbf{72.98} \\ 
DT & 69.62 & 70.60 & 71.81 & \textbf{71.85} \\ 
RF & 78.80 & 79.01 & 79.28 & \textbf{79.49} \\ 
XGBoost & 78.94 & 79.47 & 80.01 & \textbf{80.02} \\ 
\hline 
Average & 70.94 & 73.47 & \textbf{73.88} & 73.71 \\ 
\hline
\end{tabular}
\end{table}

A win-loss analysis against traditional CSL is reported in Table \ref{tab:multiclass_win_loss}. Across the 50 classifier-dataset combinations, Neighbor-iCost outperforms traditional CSL in 33 cases, loses in 13 cases, and ties in 4 cases. Gini-iCost achieves 28 wins, 17 losses, and 5 ties.

\begin{table}[!htbp]
\centering
\small
\caption{Overall win-loss comparison of iCost variants against CSL on multiclass datasets based on MCC.}
\label{tab:multiclass_win_loss}
\begin{tabular}{lcccc}
\hline
\textbf{Method} & \textbf{Avg. \(\Delta\)MCC} & \textbf{iCost Wins} & \textbf{CSL Wins} & \textbf{Ties} \\
\hline
Neighbor-iCost & 0.41 & 33 & 13 & 4 \\
Gini-iCost     & 0.24 & 28 & 17 & 5 \\
\hline
\end{tabular}
\end{table}

Multiclass classification is naturally more complex than binary classification \cite{lango2022makes}. In imbalanced multiclass problems, the difficulty increases further because several minority classes may coexist with different levels of overlap, sparsity, and class-specific ambiguity. Although the OvR strategy provides a simple and effective way to extend binary iCost to multiclass settings, it also has an inherent limitation: for each target class, all remaining classes are grouped into a single ``rest'' class. As a result, the class-specific structure among the non-target classes is not explicitly preserved during cost assignment. This may partly explain why the performance gains in the multiclass experiments are smaller than those observed in the binary experiments. Further investigation is therefore needed to develop multiclass-specific instance-complexity measures, such as pairwise one-vs-one cost modeling or direct multiclass regional complexity estimation. Overall, the present results show that iCost can be extended to multiclass imbalanced classification, while also highlighting multiclass complexity modeling as an important direction for future work.


\subsection{Discussion, Implications, and Future Directions}

From a methodological standpoint, the findings show that minority-class instances should not necessarily be treated with equal importance. The main benefit of iCost comes from relaxing the rigidity of conventional class-level cost assignment. Traditional CSL assigns the same penalty to all minority-class instances, although these samples may differ substantially in terms of local safety, overlap, boundary ambiguity, and outlier-like behavior. By assigning costs according to instance complexity, iCost provides a more flexible learning strategy that better reflects the heterogeneous nature of minority samples.

iCost can be integrated with different classifiers, without modifying the original training data distribution. This is useful because it avoids generating synthetic samples or removing existing samples, which can sometimes introduce noise or discard useful information. It also has far less computational overhead than resampling techniques, which can become computationally expensive for large datasets. The results also indicate that iCost improves the balance between FPs and FNs. The proposed method does not simply force the classifier to predict more minority samples. As a result, it achieves better performance while avoiding excessive majority-class misclassifications. This is important because, in predictive modeling, improving sensitivity alone may not indicate better overall performance if it leads to a large increase in FPs.

Several limitations remain. First, although the tuning analysis demonstrates the benefit of cost-parameter optimization, the search space was intentionally kept compact to maintain computational feasibility across multiple datasets and classifiers. A finer or adaptive cost-search strategy may further improve performance. Second, the proposed complexity estimates are based on either local neighborhood composition or Gini-impurity-based feature-space partitioning. These measures are effective and interpretable, but they may not fully capture complex nonlinear structures in high-dimensional data. Third, the multiclass extension relies on OvR decomposition. While this provides a straightforward way to apply iCost to multiclass problems, it merges all non-target classes into a single rest class and therefore does not explicitly preserve class-specific relationships among the non-target classes.

Future work can extend the proposed framework in several directions. One direction is to investigate adaptive or data-driven cost tuning. Another direction is to develop multiclass-specific complexity measures, such as pairwise one-vs-one cost modeling or direct multiclass regional ambiguity estimation. Finally, other instance-complexity estimation strategies can be explored, including density-aware, margin-based, or graph-based measures.

\section{Conclusion}

This study proposes iCost, an instance-complexity-aware cost-sensitive learning framework for imbalanced classification. Instead of assigning a uniform penalty to all minority-class samples, iCost adjusts the learning cost according to the estimated complexity of each minority instance. Two variants were developed: Neighbor-iCost, which estimates complexity from local neighborhood composition, and Gini-iCost, which uses Gini-impurity-based feature-space partitioning to identify regions of different class ambiguity.

Extensive experiments on 75 imbalanced datasets using five classifiers show that the proposed approach achieves clear overall improvements over conventional cost-sensitive learning. It is also highly competitive with widely used resampling methods while preserving the original training distribution. Since it does not generate synthetic samples or remove existing samples, iCost can be applied to large datasets with limited additional computational overhead. The framework is also directly applicable to both binary and multiclass classification settings.

By tailoring instance penalization to sample difficulty, iCost improves predictive performance and provides a more flexible alternative to class-level cost assignment. To support reproducibility and practical use, the proposed algorithm has been released as a scikit-learn-compatible Python package through PyPI. Overall, this work offers a fresh perspective on imbalanced learning and lays the groundwork for future methods that focus on the intrinsic structure of the data.

\section*{Appendix}

Supplementary files and source code are available in this repository: 

https://github.com/newaz-aa/iCost



\section*{Declaration of generative AI and AI-assisted technologies in the manuscript preparation process}

During the preparation of this work, the authors used ChatGPT by OpenAI for language editing, grammar checking, improving clarity, assisting with manuscript organization, and organizing result spreadsheets for presentation. The authors reviewed and edited the output as needed and take full responsibility for the content of the published article.


\bibliographystyle{elsarticle-num}
\bibliography{references}


\end{document}